\documentclass{article}

\usepackage[final]{corl_2026} % Uncomment for the camera-ready ``final'' version.
\usepackage{microtype}
\usepackage{amsmath, amssymb, amsthm}
\usepackage{mathtools}
\usepackage{wrapfig}
\usepackage{graphicx}
\usepackage{algorithm}
\usepackage{algorithmic}
\usepackage{booktabs}
\usepackage{enumitem}
\usepackage{pifont}
\usepackage{multirow} 
\usepackage{makecell}  
\usepackage{caption}
\usepackage{adjustbox}

\newcommand{\cmark}{\ding{51}}
\newcommand{\xmark}{\ding{55}}
\newcommand{\slot}[1]{\texttt{\textless #1\textgreater}}

\let\oldding\ding% Store old \ding in \oldding
\renewcommand{\ding}[2][1]{\scalebox{#1}{\oldding{#2}}}

\title{Cue the Flow: Steering Flow-Matching Policies \\ for Open-World Delivery Manipulation}

\author{\textbf{Haoxuan Wang}$^{1}$ \qquad 
\textbf{Griffin Galimi}$^{2}$ \qquad 
\textbf{Junhua Huang}$^{2}$ \\
\textbf{Selina Song}$^{2}$ \qquad 
\textbf{Wayne Wu}$^{2}$ \qquad 
\textbf{Yan Yan}$^{1}$ \qquad \textbf{Bolei Zhou}$^{2}$\\
$^{1}$University of Illinois Chicago \quad
$^{2}$University of California, Los Angeles \\
\url{https://hatchetproject.github.io/delivery_steer/}
}

\begin{document}
\maketitle

%===============================================================================

\begin{abstract}
    Open-world goods delivery requires mobile manipulators to follow free-form user instructions and manipulate potentially novel objects. Existing dual-system approaches use high-level grounding models to convert language into grounded visual prompts, but their low-level controllers can remain brittle under noisy perception, dynamic scenes, and contact-rich interactions. We instead use a pretrained flow-matching vision-language-action model as the low-level control interface, leveraging its reactivity and robustness to environmental changes while treating the grounding output as a spatial cue for policy steering. Our key insight is that the pretrained VLA already provides a strong manipulation prior, while the spatial cue supplies the missing target information needed to guide actions under novel language--object mappings. Concretely, we introduce a lightweight cue-conditioned adapter. The adapter is first trained with contrastive objectives to produce salient and spatially discriminative cue representations, and is then supervised to predict a diagonal affine transformation over the generated action chunk, aligning policy steering with the cued target. Across tabletop and mobile-base settings, our method improves instruction following and manipulation success on both in-domain and out-of-domain objects, achieving up to near $2\times$ improvement in average task success rate with negligible inference overhead.
\end{abstract}


%===============================================================================

\section{Introduction}
\begin{figure}[t]
    \vspace{-10pt}
    \centering
    \includegraphics[width=1.0\linewidth, trim=15 110 5 110, clip]{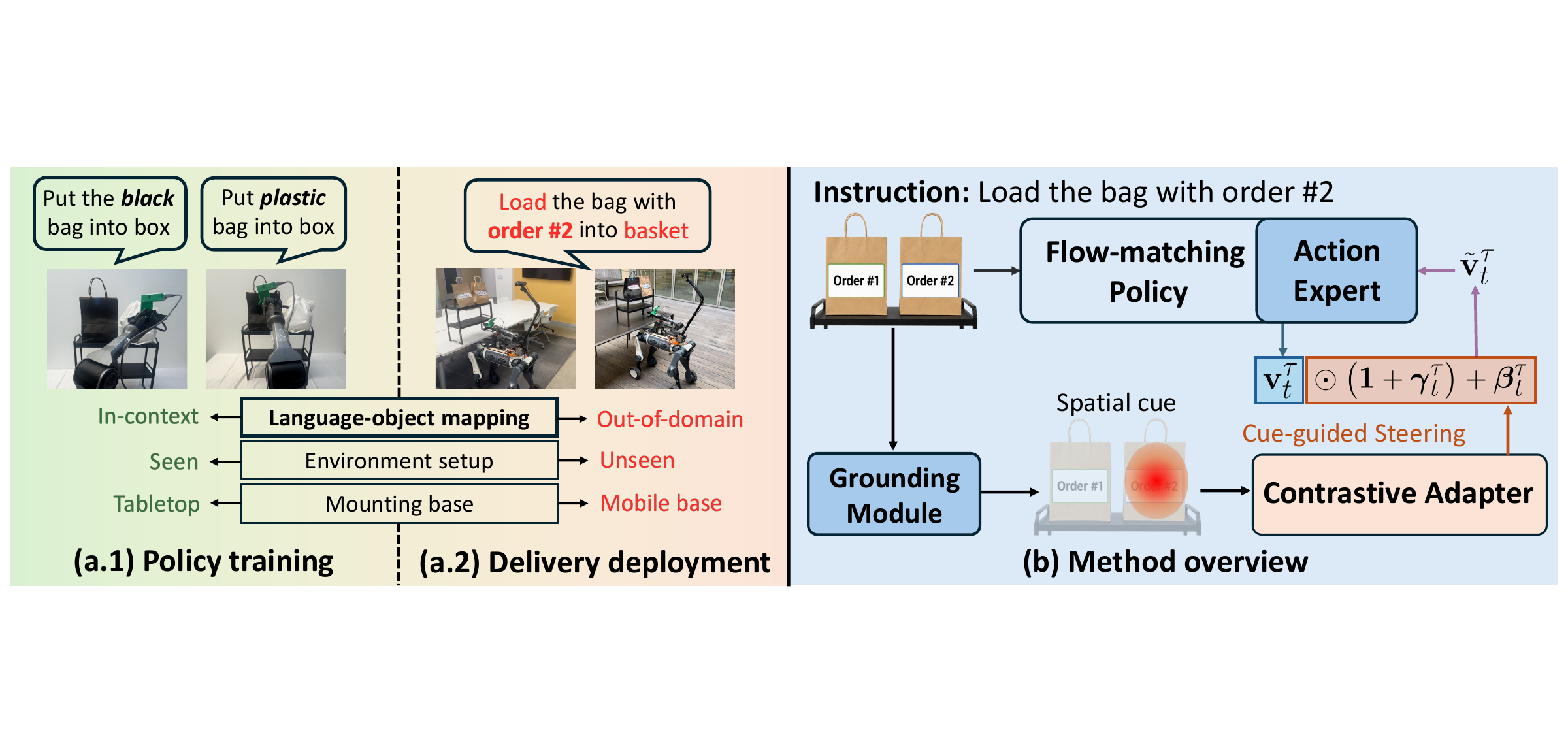}
    \vspace{-14pt}
    \caption{
    % We consider the manipulation phase of the delivery task. 
    \textbf{(a.1)} The policy is trained on in-context demonstrations collected in a fixed environment.
    \textbf{(a.2)} In real-world delivery, however, the robot must execute manipulation under changing environments and novel language--object mappings.
    \textbf{(b)} We focus on improving policy generalization to these novel mappings. A grounding module converts the open-ended instruction into a spatial cue, which conditions a contrastive adapter to predict affine transformations over the frozen flow-matching policy's action output, steering the generated action chunk toward the target object.
    % \wayne{{b} was not referred}
    }
    \label{fig:pipeline}
    \vspace{-18pt}
\end{figure}

%===============================================================================
% Successful execution requires more than a generic delivery skill. 
Consider a delivery robot receiving the request: ``Give me the takeout bag with the name Joseph on it.'' We focus on the \textbf{manipulation phase} of this task, where the robot must turn an open-ended instruction into a physical action: parsing the language, grounding the referred object in a potentially cluttered scene, and executing a reliable trajectory toward it. While the underlying manipulation skill may be simple, such as pick-and-place, learning a practical policy for open-world delivery remains difficult. As shown in Figure~\ref{fig:pipeline}(a), the core challenge is a substantial train--deployment mismatch: the policy is trained in fixed environments with a limited vocabulary of objects and instructions, yet deployed on a mobile robot in open-world settings with novel objects, unseen referring expressions, and variable scene configurations.
% In this work, we particularly focus on the  of the delivery task, where language grounding must be translated into executable manipulation behavior. 
% This setting is common in last-meter delivery, where user requests and object appearances vary across deployments, and the target may fall outside the policy's training distribution.

% \wayne{give a figure for task definition, maybe make a teaser figure, which can clearly show what the task is; in contribution, we can also say this new task is a contribution, so here we need to clearly define the task.}
Large-scale pretrained robot policies~\citep{openvla,rt2,pi0,internvla} have substantially improved policy robustness under visual perturbations~\citep{colosseum}, reducing the brittleness caused by changes in scene appearance and environment layout. Yet a central challenge remains: how to extend a policy to novel language--object mappings while preserving its previously learned manipulation capability.
One promising direction is a dual-system design that decouples explicit grounding from robot control~\citep{rekep,moka,robopoint,vp-vla}. The high-level grounding module, or System~2, interprets the instruction and converts it into spatial abstractions, often represented as visual prompts~\citep{vp_1,vp_2}. The low-level controller, or System~1, then conditions on these prompts to generate executable robot actions.
% through optimization-based motion generation.
% such as object masks, keypoints, or waypoints, 
% , often with the help of open-vocabulary grounding or segmentation models~\citep{qwen3vl,sam3,groundingdino}. A low-level System~1 controller then converts these spatial abstractions into robot actions. This decomposition makes the language-to-object grounding more explicit and reduces the burden on the controller. 

\begin{wraptable}{r}{0.5\linewidth}
    \vspace{-2pt}
    \centering
    \adjustbox{width=\linewidth}{
    \begin{tabular}{l|ccc}
    \toprule
    Paradigms 
    & VLA~\citep{pi0} 
    & \makecell{Dual +\\ low-level~\citep{moka}} 
    & \makecell{\textbf{Dual +}\\ \textbf{VLA (ours})} \\ 
    \midrule
    Learned control      & \cmark    & \xmark    & \cmark \\
    Precise perception   & \cmark    & \xmark    & \cmark \\
    Explicit grounding   & \xmark    & \cmark    & \cmark \\
    OOD generalization   & Limited   & Strong    & Strong \\
    \bottomrule
    \end{tabular}
    }
    \vspace{-6pt}
    \caption{\textbf{Comparison of paradigm designs.}}
    \label{tab:paradigm_comparison}
    \vspace{-14pt}
\end{wraptable}
However, many existing dual-system methods still depend on optimization-based or geometric motion generation~\citep{kpam,moka,tracevla}. Such controllers can be fragile in open-world delivery settings: perception may be noisy, the target object may move during execution, and a fixed plan may not recover from grounding or localization errors. This motivates replacing the geometric low-level controller with a pretrained robot policy, especially a flow-matching vision-language-action model~\citep{pi05,gr00t,smolvla}. Since these models are trained on diverse robotic demonstrations, they provide learned priors for object interaction, contact-rich manipulation, and reactive correction. They therefore offer a practical low-level control interface for the dual-system design, while preserving the explicit grounding ability of the high-level module. Table~\ref{tab:paradigm_comparison} compares these paradigms.
% \wayne{This paragraph is more like analysis for general manipulation; we should start from our specific application. What is the challenge for the delivery application? Why can current models not handle it well? Then comes flow-matching, and then comes our method. Should center around our new application to state.}

% These limitations motivate using large-scale pretrained robot policies~\citep{openvla,smolvla,pi0,internvla} as learned System~1 controllers~\citep{vp-vla}. Unlike hand-designed motion modules, pretrained VLAs can encode implicit dynamics, contact-rich behaviors, and reactive corrections through training on diverse robot demonstration data. 
% Among them, flow-matching vision-language-action models~\citep{pi05,gr00t,smolvla} are particularly attractive because they generate expressive action chunks through an efficient continuous generation process.

Yet integrating flow-matching VLAs into a dual-system design is not straightforward. The visual prompt produced by System~2 must be converted into a form that can reliably steer a policy trained to act from its own visual-language representations. Prior work~\citep{vp-vla,vca} explores full policy training with visual prompts as additional inputs. While effective in controlled settings, this strategy still depends on the VLA's internal spatial reasoning ability, which is known to be limited~\citep{spatialvla,vipa}. More critically, input-level prompting does not specify how the prompt should modulate the action generation process. After being encoded with the rest of the visual observation, the prompt is only implicitly represented, without a direct mechanism that encourages the generated trajectory to align with the System~2 guidance. The key challenge is therefore not merely how to provide a visual prompt to a VLA, but how to turn that grounded signal into actionable guidance during flow-based action generation.

In this paper, we address this challenge by treating the grounding output as a spatial cue that directly guides action generation. Our key insight is that a pretrained flow-matching VLA already knows how to manipulate: it can generate feasible trajectories for objects in the workspace, but may not reliably determine which object to act on under unseen language--object mappings. Thus, the missing component is not a new manipulation policy, but a mechanism for specifying the intended target.
We therefore keep the VLA frozen and introduce a lightweight cue-conditioned adapter that uses the spatial cue generated by System~2 to steer the frozen policy's action output. The adapter first learns to encode the explicit cue into a salient and spatially discriminative representation, using contrastive objectives to discourage reliance on scene-specific shortcuts. It then conditions on this cue representation to predict a diagonal affine transformation over the generated action chunk, shifting the frozen policy's output toward the cued object while preserving its learned manipulation prior. Figure~\ref{fig:pipeline}(b) provides an overview of the full pipeline.

% The adapter is trained to predict a diagonal affine transformation over the frozen policy's action output. To prevent the adapter from exploiting shortcut correlations in scene-specific visual patterns, rather than learning the intended cue--action correspondence, we train it with contrastive objectives that make the cue representation both spatially discriminative and salient. 
% This preserves the base policy's manipulation prior while enabling better generalization to novel objects and instructions.
% These objectives encourage the learned transformation to follow the spatial cue, rather than incidental background features.
% \wayne{conclude the contribution briefly}

Our contributions are threefold. First, we analyze the policy-learning challenges posed by open-world delivery and propose a dual-system framework that combines explicit language--object grounding with learned low-level action generation. Second, we develop a lightweight cue-conditioned adapter for frozen flow-matching VLAs that steers generated action chunks via a diagonal affine transformation while avoiding full policy fine-tuning. Third, we validate the proposed method across three in-domain and out-of-domain object settings. Our method consistently outperforms all baselines, achieving relative gains of 86\% and 44\% over the best-performing baselines in average task success rate under tabletop and mobile-base settings, respectively.

\section{Related Work}
\textbf{Flow-matching VLAs.}
A central goal in robot learning is to build generalist policies that can perform diverse tasks and transfer across environments, objects, and embodiments. Vision-language-action (VLA) models~\citep{openvla,rt2,gr00t,pi05} address this goal by mapping visual observations and language instructions directly to robot actions, often after training on large-scale robot and human demonstration data. Recent flow-matching VLAs~\citep{gr00t,pi05,smolvla} further model action generation as a learned continuous flow conditioned on fused visual-language features, allowing them to represent complex and potentially multimodal action distributions~\citep{ado_noising}. Although flow-matching VLAs have begun to be used within dual-system architectures~\citep{hirobot,pointvla}, how to effectively steer their action generation toward novel language--object mappings remains underexplored.

\textbf{Policy steering.}
Policy steering improves pretrained generative policies by modifying their sampling or refinement process while keeping the base policy fixed~\citep{steer_rl}. Existing methods typically guide generated actions with external objectives or auxiliary models, steering outputs toward task, safety, or user-specified constraints. Model-predictive refinement methods~\citep{mpc_safety,omniguide} use learned dynamics models to improve task performance or enforce safety, while human-in-the-loop approaches~\citep{human_loop_1,human_loop_2,human_loop_3} refine actions using user-provided subgoals, corrections, or preferences. Classifier- and dynamics-guided methods~\citep{dynaguide,lpb} further leverage latent visual dynamics models~\citep{dynamics_model,daydreamer} to bias actions toward desired outcomes. Unlike prior steering methods that primarily use external signals to refine in-domain behavior, our method treats spatial cues as target-specification signals and learns cue-conditioned transformations over frozen VLA action chunks, enabling better generalization to out-of-domain language--object mappings.

\textbf{Visual prompting for robotic policies.}
Explicit visual representations provide a useful interface for guiding robotic manipulation. One line of work uses visual marks to connect a vision-language model (VLM) planner with a low-level controller~\citep{moka,rekep,robopoint}, where the controller translates visual predictions into executable actions. However, such controllers can be brittle in open-world settings where perception is noisy and reactive control is needed. Another line of work treats actions as language~\citep{crayonrobo,action_language}, using the VLM to directly output executable robot actions, but these methods depend heavily on the grounding precision and reasoning reliability of the VLM. Other approaches~\citep{tracevla,vp-vla,vap} inject visual prompts into the VLA input stream, yet the prompt's influence on action generation remains implicit. In contrast, our method uses a spatial cue separated from the input image as an explicit target-specification signal, steering a frozen VLA by directly modulating its generated action chunk.
% largely keep the underlying behavior-cloning backbone unchanged and rely on end-to-end imitation from hindsight-labeled demonstrations to learn the corresponding motion. 
% These studies establish visual prompting as a lightweight interface between perception and action, without requiring a redesign of the underlying VLA. 

%===============================================================================

% \subsection{Preliminaries}
% \paragraph{Flow-matching VLA.}
% Flow-matching policies~\citep{flow_matching} generate actions by learning a continuous-time velocity field that transports samples from a simple prior, such as Gaussian noise, to the action distribution. In flow-matching VLAs, the current observation context is first encoded by the VLM backbone into a conditional representation $\mathcal{K}$. The action expert then conditions on $\mathcal{K}$ and transforms an initial noise sample $\mathbf{x}_0 \sim \mathcal{N}(\mathbf{0}, \mathbf{I})$ into an action chunk by integrating the learned velocity field $\mathbf{v}_\pi$ over normalized flow time $s \in [0,1]$. With $N$ Euler integration steps, the update is
% \begin{equation}
%     \mathbf{x}_{s + \frac{1}{N}}=\mathbf{x}_s + \frac{1}{N} \mathbf{v}_\pi(\mathbf{x}_s, s \mid \mathcal{K}),
%     \label{eq:euler}
% \end{equation}
% where integration starts from $\mathbf{x}_0$ and terminates at $\mathbf{x}_1$. The final output $\mathbf{x}_1 \in \mathbb{R}^{H \times D}$ is an action chunk of horizon $H$, with each action having dimension $D$.
% \td{maybe add a plot to show the problem, since it is a contribution}
\section{Method}
We present the problem setup and detail method design in this section. Section~\ref{sec:prob} formulates generalization to novel language--object mappings as a policy-steering problem. Section~\ref{sec:sys_design} introduces the overall dual-system framework. Section~\ref{sec:adapter_learning} describes the spatial-cue-conditioned adapter for steering the frozen policy. Section~\ref{sec:impl_details} provides the implementation details.

\subsection{Problem Definition}
\label{sec:prob}
Consider two deployments of the same frozen flow-matching policy. In the first, both the instruction and the referred object fall within the training distribution, and the policy can execute the task successfully. In the second, the target object is replaced by a novel object and the instruction is changed accordingly. The underlying motor behavior may remain the same, such as reaching, grasping, and placing, but the policy must now associate the instruction and visual scene with an unseen target.
The central challenge is therefore not to acquire a new manipulation skill, but to redirect an existing manipulation prior toward the intended object during action generation. We formulate this as a policy-steering problem: given a frozen flow-matching VLA and an explicit spatial cue produced by a grounding module, the goal is to modulate the generated action chunk so that it follows the cued target while preserving the base policy's learned manipulation capability.
% and the policy can execute the corresponding manipulation behavior. 

% Let $\mathbf{o}_t$ denote the observation-language input at time $t$, and 
% Let $\mathbf{A}_t$ denote the predicted action chunk at time $t$. 
% A flow-matching policy generates action chunks by integrating a learned velocity field.
% Our objective is to preserve the policy's in-domain behavior while improving its generalization to novel objects by steering the pretrained policy through a spatial cue $\mathbf{h}_t$ generated by the System~2 planner.
% By utilizing the heatmap cue $\mathbf{h}_t$, the new policy can be steered to operate on the novel objects.

% We consider a \textbf{flow-matching policy} $\mathbf{v}{_\pi}(\mathbf{A}_{t} | \mathbf{o}_{t})$ that is pretrained on large-scaled cross-modality data and finetuned on limited task-specific data, where $\mathbf{o}_t$ denote the input observation and language prompts, and $\mathbf{A}_t$ as the output action chunk. 
% Our objective is to improve its zero-shot generalizability by learning a spatial cue guided policy $\hat{\mathbf{v}}_\pi(\mathbf{A}_{t} | \mathbf{o}_{t},\mathbf{h}_t)$, built upon $\mathbf{v}{_\pi}(\mathbf{A}_{t} | \mathbf{o}_{t})$. By utilizing the heatmap cue $\mathbf{h}_t$, the new policy can be steered to operate on the novel objects.

\subsection{Dual-System Design}
\label{sec:sys_design}
% \wayne{figure \label{fig:pipeline} (b) should be referred in this section? if yes, that figure is too far away, may be moved to this section}
Flow-matching VLAs generate action chunks using an action expert conditioned on features from a vision-language backbone. While this end-to-end grounding is effective for in-distribution tasks, it can become brittle when the policy encounters novel instructions referring to unseen objects. To reduce this uncertainty, we adopt a dual-system design that separates language--object grounding from action generation. The high-level grounding module converts diverse free-form instructions into a unified spatial cue, allowing the downstream policy to be steered by explicit target information rather than relying solely on implicit language grounding.
% A perception tier, referred to as System 2, performs open-vocabulary language-conditioned visual grounding and produces a structured spatial cue for the target object. The VLA policy, referred to as System 1, then uses this cue to generate the action chunk. This design reduces the grounding burden on the frozen VLA while preserving its pretrained manipulation skills, allowing the policy to focus its action generation on the cued target rather than relying solely on implicit feature alignment.

\textbf{System~2: Grounding module.}
System~2 converts open-ended user instructions into a structured \textbf{spatial cue} for the downstream VLA, specifying the 2D target location in the workspace camera view. We first use a grounding model to predict spatial references for the target object, typically in the form of bounding boxes. An external segmentation model~\citep{sam2,sam3} then converts these references into a pixel-level target mask. From this mask, we compute the object centroid and construct a Gaussian heatmap $\mathbf{h}_t$. This heatmap is not injected as an image-level visual prompt but used as an explicit spatial cue. All components in System~2 are frozen and used for zero-shot inference.
% so the module provides explicit grounding information without modifying the pretrained perception or action models.
% It also rewrites the user instruction into a standardized command format compatible with the VLA's training set, removing task-irrelevant information while unifying the input format. 

\textbf{System~1: VLA as controller interface.}
To preserve the pretrained manipulation capability of the VLA while enabling it to be steered by the spatial cue from System~2, we freeze the VLA policy and attach a lightweight adapter. The adapter predicts a cue-conditioned diagonal affine transformation that modulates the action chunk produced by the frozen policy during flow integration.
Specifically, let $\mathbf{A}^{\tau}_t \in \mathbb{R}^{H \times d_a}$ denote the intermediate action chunk at integration time $\tau \in [0,1]$, and let $\mathbf{v}^{\tau}_t$ be the velocity predicted by the frozen action expert. Rather than editing the chunk directly, the adapter steers the \emph{generative path}: at each integration step it predicts two modulation terms $\boldsymbol{\gamma}^{\tau}_t, \boldsymbol{\beta}^{\tau}_t \in \mathbb{R}^{d_a}$ and applies them to the predicted velocity
\begin{equation}
\tilde{\mathbf{v}}^{\tau}_t
\;=\;
\mathbf{v}^{\tau}_t
\odot
\big(\mathbf{1} + \boldsymbol{\gamma}^{\tau}_t\big)
\;+\;
\boldsymbol{\beta}^{\tau}_t ,
\label{eq:steer}
\end{equation}
after which integration continues with the steered velocity and $\mathbf{A}^{\tau+\Delta\tau}_t = \mathbf{A}^{\tau}_t + \Delta\tau\,\tilde{\mathbf{v}}^{\tau}_t$, where integration proceeds from the noise sample at $\tau{=}1$ to the executed chunk at $\tau{=}0$ with signed step $\Delta\tau = -1/N$. Here $\odot$ denotes element-wise multiplication, and $\boldsymbol{\gamma}^{\tau}_t,\boldsymbol{\beta}^{\tau}_t$ are broadcast across all $H$ positions of the chunk. The transformation is \emph{diagonal} in the action dimensions, where the scaling term $\boldsymbol{\gamma}^{\tau}_t$ reweights the frozen policy's predicted velocity, while the shift term $\boldsymbol{\beta}^{\tau}_t$ introduces a cue-dependent drift. Because the modulation is re-predicted at every integration step, the frozen policy re-evaluates at the steered state and continues to shape the trajectory, so the generated chunk is redirected toward the cued target while remaining governed by the pretrained flow.

% The adapter is not intended to synthesize new motor behaviors from scratch. Instead, it biases the pretrained policy toward an appropriate trajectory mode already supported by the base action distribution. 
% This design preserves the manipulation capability of the VLA while allowing the policy to shift its generated action chunk toward the visually cued target.
% Weaken the relationship between raw image input and the action trajectory, while enhancing the relationship between spatial cue and the action trajectory. But the spatial cue and raw image input are always binded, to , we therefore introduce the contrastive learning strategy.
% However, the spatial cues, though being intuitive to human perception, often do not provide significant guidance signals when applied to flow-matching VLAs. This is because these spatial cues are added onto the input images and then encoded into latent features. Without abundant data that trains the VLM backbone to effectively focus on such cues, the guidance by such spatial cues can be easily ignored.

\subsection{Contrastive Adapter}
\label{sec:adapter_learning}
\begin{wrapfigure}{r}{0.5\textwidth}
    \vspace{-24pt}
    \centering
    \includegraphics[width=1.0\linewidth, trim=250 100 430 170, clip]{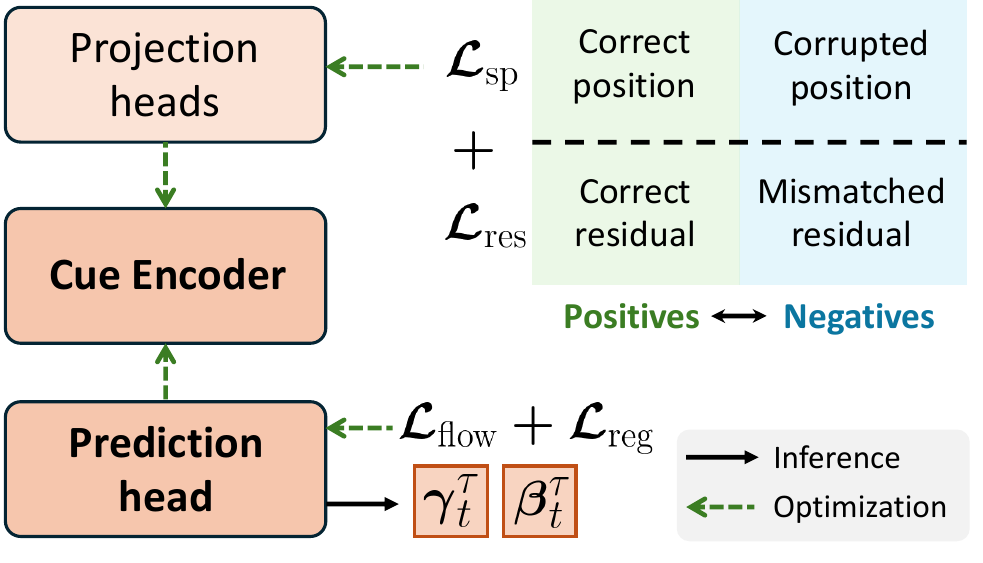}
    \vspace{-16pt}
    \caption{\textbf{Illustration of adapter learning.}}
    \label{fig:adapter}
    \vspace{-10pt}
\end{wrapfigure}
Learning an effective adapter is challenging because the spatial cue is aligned with the target object in the observation. Without additional constraints, the adapter may exploit shortcut correlations from in-domain training scenes or memorize scene-specific visual patterns, rather than using the cue as the source of target specification. To strengthen the correspondence between cue location and action transformation, we design the adapter with two components: a cue encoder that learns salient and spatially discriminative features from the explicit cue, and a prediction head that maps these features to affine transformation parameters for steering the frozen policy output. Figure~\ref{fig:adapter} illustrates the adapter architecture and training pipeline.
% The two modules are learned as follows.

\textbf{Cue encoder learning.}
The spatial cue specifies the target location but does not by itself encode object semantics or task context. We therefore train a cue encoder $\mathcal{E}$ to transform the cue $\mathbf{h}_t$ into a representation suitable for action steering. The learned representation is encouraged to satisfy two properties: \textbf{spatial discrimination}, preserving where the target lies relative to the action representation, and \textbf{cue salience}, emphasizing cue-relevant features over incidental visual patterns.
% We implement these two objectives through contrastive learning based on InfoNCE~\citep{infonce}.
 
Let $\mathcal{E}$ denote the cue encoder. To make the cue representation spatially discriminative, we require the action representation to align more closely with its corresponding spatial cue than with spatially shifted versions of the same cue. We first encode the Gaussian heatmap and project it into a cue representation,
$
\mathbf{z}_c = \mathcal{P}_c(\mathcal{E}(\mathbf{h}_t)),
$
where $\mathcal{P}_c$ is a projection head. We also construct an action representation, $\mathbf{z}_a = \mathcal{P}_a(\mathbf{a}_t, \mathcal{E}(\mathbf{h}_t)),$
where $\mathbf{a}_t$ denotes the action-expert feature and $\mathcal{P}_a$ is the corresponding projection head.
For each positive cue representation $\mathbf{z}_c^i$, we construct a set of \emph{spatial-shift negatives} $\{\mathbf{z}_{c,k}^{-}\}_{k=1}^{K}$ by perturbing the heatmap peak on the patch grid while keeping the underlying scene fixed. These negatives retain the same visual context but specify different target locations, forcing the cue representation to capture spatial position rather than appearance context. The spatial contrastive loss for sample $i$ is
\begin{equation}
\mathcal{L}_{\text{sp}}^{i}
=
-\log
\frac{
\exp(\mathbf{z}_a^i \cdot \mathbf{z}_c^i / \tau)
}{
\sum_{j=1}^{B}
\exp(\mathbf{z}_a^i \cdot \mathbf{z}_c^j / \tau)
+
\sum_{k=1}^{K}
\exp(\mathbf{z}_a^i \cdot \mathbf{z}_{c,k}^{-} / \tau)
},
\label{eq:cab}
\end{equation}
where $B$ is the batch size and $\tau$ is the softmax temperature. This objective contrasts the correct cue location against both in-batch cues and spatially corrupted cues, encouraging the encoder to learn features that are sensitive to the target location rather than scene-level appearance.
% Specifically, given the per-sample heatmaps with peaks at location $(u, v)$, we sample each shift uniformly from $\Delta \in \{\pm 3, \ldots, \pm 7\}^{2}$ patches and apply a zero-padded 2D translation:
% \begin{equation}
% \tilde h^{k}_{u,v} \;=\;
% \begin{cases}
% h_{u-\Delta_u^{k},\, v-\Delta_v^{k}}
%   & \text{if } (u-\Delta_u^{k}, v-\Delta_v^{k}) \in [0,H)\!\times\![0,W), \\ 
% 0 & \text{otherwise,}
% \end{cases}
% \end{equation}
% where $H$ and $W$ are the height and width of the encoded image's patch grid. Each shifted heatmap pair is passed through the shared cue encoder
% and projection to produce a negative sample $z_{c,k}^{-}$. And we thus adopt the loss~\citep{infonce} as:

Since the adapter also conditions on action-related features, it may learn shortcuts that are not driven by the spatial cue itself. We therefore introduce a residual cue representation that isolates the effect of the cue. For each sample, we compute two encodings: one from the real heatmap, $\mathcal{E}(\mathbf{h}_t)$, and one from a zero heatmap, $\mathcal{E}(\mathbf{0})$. Their difference captures the cue-induced residual, which is then projected into the same representation space as the action feature:
$
\mathbf{z}_{\text{res}}
=
\mathcal{P}_\delta\big(\mathcal{E}(\mathbf{h}_t) - \mathcal{E}(\mathbf{0})\big),
$
where $\mathcal{P}_\delta$ is a projection head. This reinforce the adapter to learn cue-relevant action transformations. Therefore, for each sample $i$, we introduce the residual contrastive loss:
\begin{equation}
\mathcal{L}_{\text{res}}^{i}
\;=\;
-
\log
\frac{
\exp(\mathbf{z}_a^{i}\!\cdot \mathbf{z}_{\text{res}}^{i}/\tau)
}{
\sum_{j=1}^{B}
\exp(\mathbf{z}_a^{i}\!\cdot \mathbf{z}_{\text{res}}^{j}/\tau)
}.
\label{eq:rab}
\end{equation}
% This objective encourages the encoder to preserve information that is specifically introduced by the spatial cue and predictive of the target-directed action.

\paragraph{Affine transformation learning.}
% The contrastive objectives in Eqs.~\ref{eq:cab} and~\ref{eq:rab} encourage the cue encoder $\mathcal{E}$ to produce representations that are both spatially discriminative and aligned with target-directed action features. 
Given the encoded cue, a prediction head outputs the modulation terms $\boldsymbol{\gamma}^{\tau}_t$ and $\boldsymbol{\beta}^{\tau}_t$, which define the diagonal affine transformation in Eq.~\eqref{eq:steer}. We train this prediction head under the same flow-matching supervision as the base policy, so that the adapter learns to steer the frozen policy's action chunks while preserving the pretrained action distribution.
To discourage overly large corrections, we regularize the magnitude of the modulation terms:
\begin{equation}
\mathcal{L}_{\text{reg}}
\;=\;
   \|\boldsymbol{\gamma}^{\tau}_t\|_2^{2}
   +
   \|\boldsymbol{\beta}^{\tau}_t\|_2^{2}.
\label{eq:reg}
\end{equation}

The full training objective is therefore
\begin{equation}
\mathcal{L}
\;=\;
\mathcal{L}_{\text{flow}}
\;+\;
\lambda_{\text{sp}}\,\mathcal{L}_{\text{sp}}
\;+\;
\lambda_{\text{res}}\,\mathcal{L}_{\text{res}}
\;+\;
\lambda_{\text{reg}}\,\mathcal{L}_{\text{reg}},
\label{eq:total_loss}
\end{equation}
where $\mathcal{L}_{\text{flow}}$ trains the affine transformation under the base flow-matching supervision, while the auxiliary terms align the transformation with the spatial cue and regularize its magnitude to avoid corrupting the pretrained action distribution. 
% We fix $\lambda_{\text{sp}}=\lambda_{\text{res}}=0.05$ and $\lambda_{\text{reg}}=0.01$ across all experiments.

% \paragraph{Termination-State Learning.}
% The adapter is trained on segments from in-domain manipulation trajectories. When deployed on novel objects, however, the terminal state may not be well represented in the training data, such that after the object has been placed, the scene contains a new post-placement configuration that the policy has not explicitly observed. In such cases, the robot may continue producing pick-and-place motions even after the task has been completed, because the scene still contains familiar manipulation cues.

% We address this issue by using cue absence as an explicit termination signal. For each training episode, we append a short synthetic segment constructed from the ending frames of the trajectory and replace the corresponding heatmap with a zero heatmap. The zero cue is paired with a no-motion or home-pose action target, so that the policy learns that the absence of a spatial cue should correspond to task completion rather than continued manipulation. 
% This construction encourages the policy to condition termination on cue absence and robot state, instead of relying on the visual identity of objects remaining in the workspace. As a result, the learned behavior can generalize more reliably across post-placement scenes, regardless of which object was just manipulated.

\subsection{Implementation Details}
\label{sec:impl_details}
% \textbf{Adapter Design.}
% The adapter is composed of several MLPs, which adds approximately $0.02\%$ trainable parameters relative to the base policy. 
% Rather than fully fine-tuning the VLA, we treat adaptation as a cue-conditioned adjustment from the original policy. The pretrained VLA, including the action expert, is kept frozen, and gradients are allowed to update only the adapter. This prevents the adaptation process from overwriting the base model's manipulation skills. Empirically, we found that full fine-tuning degrades performance significantly, motivating the use of a small sidecar adapter.
% $\mathbf{v}_{\pi}(\mathbf{A}_t \mid \mathbf{o}_t)$,
% to the visually guided policy,
% $\hat{\mathbf{v}}_{\pi}(\mathbf{A}_t \mid \mathbf{o}_t, \mathbf{h}_t)$.
% The adapter therefore uses the spatial cue from the high-level grounding module to bias the frozen policy toward an appropriate trajectory mode already present in its learned action distribution.

\textbf{Training.}
We train the adapter on the same demonstrations used to learn the flow-matching policy, supplemented with ground-truth spatial cues for target-location supervision. The adapter is implemented as a lightweight sidecar to the action expert, modulating generated action chunks while keeping the base policy frozen.
We also evaluated LoRA~\citep{lora} on the action expert as an alternative for learning cue-conditioned action transformations. In our low-data setting, LoRA tended to overfit to the training cue--action pairs, likely because its updates are directly coupled to the action-expert representations, and did not yield robust cue-steering behavior.

% we independently mask each visual token from the camera observation with a fixed probability, while keeping the heatmap input unmasked. This makes the spatial cue the most reliable target-location signal when observation tokens are corrupted, encouraging the adapter to use the cue rather than relying only on appearance features from the base observation.

% Also, for a subset of samples, we replace the original heatmap-action pair of sample $i$ with the pair from its nearest neighbor $j$ in the same batch, while keeping the base observation of sample $i$ unchanged. 
% This creates a controlled mismatch: the raw observation still corresponds to sample $i$, but the training target and spatial cue correspond to sample $j$. If the adapted policy continues to rely primarily on the raw camera input, its prediction will remain biased toward the original action of sample $i$; however, the flow-matching loss requires the trajectory associated with sample $j$. 
% This encourages the adapter to resolve the mismatch by following the paired spatial cue rather than rely primarily on the raw camera input, thereby improving cue-conditioned action steering.

\textbf{Deployment.}
During deployment, the adapter uses the same residual cue formulation as in training: it takes both the System~2 heatmap $\mathbf{h}_t$ and a zero heatmap $\mathbf{0}$, and predicts the modulation terms from their residual cue feature. For efficiency, the adapter adds about $0.02\%$ parameters relative to the base policy and introduces negligible latency. System~2 is also invoked once at the beginning of each task rollout, unless substantial scene changes require the spatial cue to be recomputed.
% we follow a server-client setup similar to \citet{pi05}, where action chunks are predicted on a remote inference server and transmitted to the robot through gRPC. 
% For termination, the cue is set to zero once task completion is detected. This decision can be provided by the upstream grounding module or by a simple robot-state criterion, such as the gripper state. No additional learned termination module or auxiliary loss is introduced.
% ; termination emerges from the same flow-matching objective applied to mildly extended training trajectories with zero-cue terminal segments.

%===============================================================================
\begin{table}[t]
\centering
\setlength{\tabcolsep}{4pt}
\renewcommand{\arraystretch}{1.15}
\adjustbox{width=1.0\linewidth}{
\begin{tabular}{lcccccccc}
\toprule
\multirow{2}{*}{\textbf{Method}} & \multicolumn{4}{c}{\textbf{Tabletop}} & \multicolumn{4}{c}{\textbf{Mobile Base}} \\
\cmidrule(lr){2-5} \cmidrule(lr){6-9}
 & Setting I & Setting II & Setting III & Average & Setting I & Setting II & Setting III & Average \\
\midrule
Base~\citep{pi05} & 70.8 & 45.8 & 0.0 & 38.9 & 33.3 & 50.0 & 8.3 & 30.5 \\
Base-L~\citep{pi05} & 37.5 & 8.3  & 29.2 & 25.0 & 8.3 & 0.0  & 8.3 & 5.5 \\
MOKA~\citep{moka} & 20.8 & 25.0 & 8.3 & 18.0 & 16.7 & 8.3 & 0.0 & 8.3 \\
VP-VLA~\citep{vp-vla} & 58.3 & 45.8 & 4.2 & 36.1 & 58.3 & 66.7 & 8.3 & 44.4 \\
\midrule
\textbf{Ours} & \textbf{83.3} & \textbf{66.7} & \textbf{66.7} & \textbf{72.2} & \textbf{83.3} & \textbf{66.7} & \textbf{41.7} & \textbf{63.9} \\
\bottomrule
\end{tabular}
}
\vspace{2pt}
\caption{\textbf{Success rates (\%) comparison under tabletop and mobile-base conditions across three settings.} (I) seen objects with paraphrased instructions, (II) unseen objects with in-domain instructions, and (III) unseen objects with novel instructions. Best results in each column are \textbf{bolded}.}
\vspace{-24pt}
\label{tab:results}
\end{table}

% We evaluate the method under both in-domain and out-of-domain real-world settings. Specifically, we aim to answer the following questions: (1) How well does our method preserve in-domain manipulation performance; (2) How well does our method generalize to novel objects and environment setups; and (3) How effective is each component design. 
\section{Experiments}
\subsection{Experimental Setup}
\textbf{Pipeline.}
For policy adaptation, we collect 80 demonstration trajectories using two training objects, a black bag and a plastic bag, without any label tags. We fine-tune $\pi_{0.5}$~\citep{pi05} on this dataset with a standard training recipe and use the resulting model as the base policy for our method. During deployment, we use Qwen3-VL-2B~\citep{qwen3vl} and SAM2.1~\citep{sam2,sam3} as the grounding module. These model choices allow the full pipeline to run simultaneously on a single RTX 5080 GPU, with the VLA operating continuously at 15 Hz. Experiments are conducted with a 6-DoF robotic arm and a wheeled dog robot, which provides vibration robustness and flexible hardware integration. No human intervention occurs during the manipulation phase we focus on, where the base remains \emph{stationary} at a fixed, predefined pose while the arm operates autonomously.

% The spatial cues used for training the adapter is generated from Qwen3-VL and SAM.
% At inference time, the System~2 components are executed sparsely because of their relatively high latency, providing visual guidance only at critical state transitions, such as the start of an episode or the onset of physical interaction. The VLA then runs continuously at 15\,Hz using the latest spatial cue.

\textbf{Evaluation.}
We evaluate under three settings. \textbf{I. Seen objects with paraphrased language} uses objects from the fine-tuning dataset with perturbed prompts. \textbf{II. Unseen objects with in-domain language} uses target objects absent from the collected trajectories, while keeping object names within the pretrained $\pi_{0.5}$ vocabulary, such as ``red bag''. \textbf{III. Unseen objects with novel language} further introduces novel referring expressions, such as ``bag with order number 1'', representing the most challenging setting.
For each task, we curate six prompts and two spatial configurations. Each prompt is generated from a compositional template, such as ``\slot{grasp-action} \slot{target-object} and \slot{place-action} into the box,'' where the grasping action, target object, and placing action vary across trials. The placement location is fixed, since delivery tasks can use a deterministic handover or drop-off location. Because all training demonstrations are collected on tabletop, deployment on the mobile base introduces an additional shift in viewpoint and scene geometry. Task success is measured as a binary outcome indicating whether the robot correctly completes the user instruction. The full set of evaluation tasks, prompts, and spatial configurations is provided in the supplementary material.

\textbf{Baselines.}
We compare our method against the following baselines:
\begin{itemize}[leftmargin=10pt,topsep=-4pt,itemsep=1pt,partopsep=1pt,parsep=1pt]
    \item \textbf{Base}~\citep{pi05}. We fine-tune pretrained $\pi_{0.5}$ on our demonstrations using one fixed language prompt per task, corresponding to a standard single-VLA setting.
    \item \textbf{Base-L}. We fine-tune pretrained $\pi_{0.5}$ with multiple language prompts per trajectory, using the same paraphrased prompts as in Setting~I. This aligns the training prompt distribution more closely with the evaluation prompts and tests whether language augmentation improves generalization.
    \item \textbf{MOKA}~\citep{moka}. MOKA is a dual-system approach with Qwen3-VL-2B and SAM2.1 as the high-level planner and a geometric low-level controller. The planner predicts 2D keypoints and waypoints, which are deprojected into 3D space for execution.
    \item \textbf{VP-VLA}~\citep{vp-vla}. VP-VLA is a dual-system approach using $\pi_{0.5}$ as System~1 and Qwen3-VL-2B with SAM2.1 as System~2. It feeds generated visual prompts directly into the policy and trains the VLA with an auxiliary grounding objective for prompted-region alignment.
\end{itemize}

\begin{figure}[t]
    \centering
    % \vspace{-6pt}
    \includegraphics[width=1.0\linewidth, trim=90 185 160 110, clip]{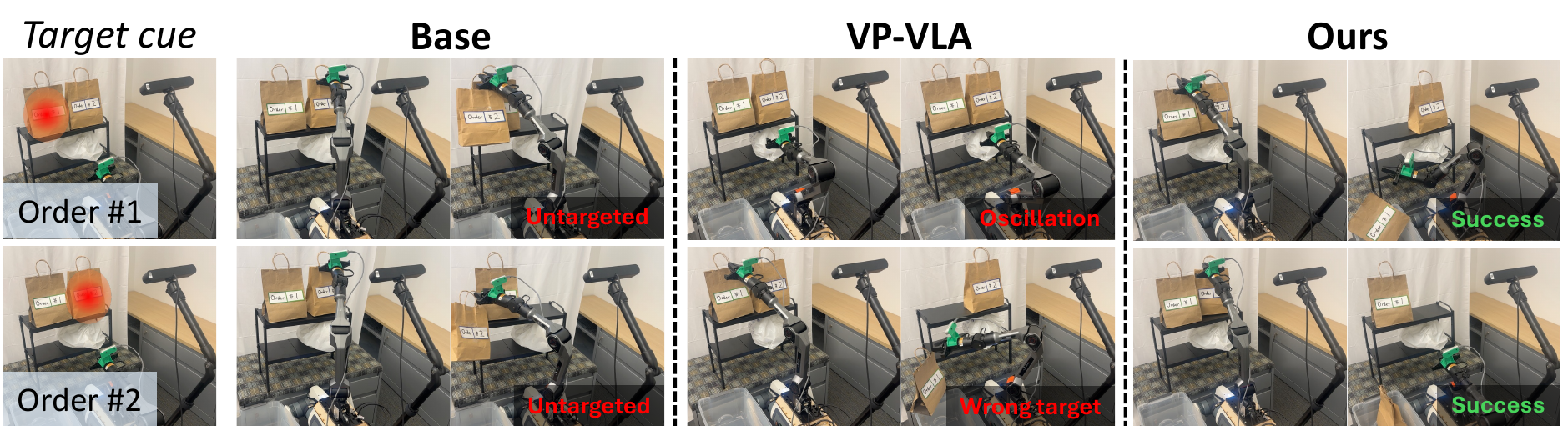}
    \vspace{-12pt}
    \caption{\textbf{Qualitative rollouts under the unseen-object with novel-language setting (Setting~III).} The leftmost column shows the System~2 spatial cue, where the red heatmap indicates the target object. For each method, two temporal keyframes are shown. The Base policy produces untargeted behavior, while VP-VLA either oscillates or reaches the wrong object. Our method follows the spatial cue and successfully manipulates the specified target in both task examples.
    % \wayne{the words (such as untargeted) may not need to be that big, but give a bbox to highlight which part is the problem shown}
    }
    \label{fig:comparison}
    \vspace{-12pt}
\end{figure}

% In these trials, the robot often produces unfocused or target-agnostic motions, suggesting that the policy fails to reliably map the novel language--object correspondence to the appropriate action distribution. This supports our motivation that the main limitation is not necessarily the absence of a manipulation skill, but the failure to ground a novel target specification during action generation.
\subsection{Results}
\textbf{Quantitative results.} Table~\ref{tab:results} summarizes the quantitative results. The mobile-base setting is generally more challenging, although the effect is not uniform across methods. Average success decreases from 38.9\% to 30.5\% for Base, from 25.0\% to 5.5\% for Base-L, from 18.0\% to 8.3\% for MOKA, and from 72.2\% to 63.9\% for our method. We infer this is due to changes in viewpoint and base instability. 
The Base policy performs strongly in Setting~I on Tabletop (70.8\%) and retains moderate performance in Setting~II (45.8\%), but fails completely in Setting~III (0.0\%). On Mobile Base, it achieves 33.3\%, 50.0\%, and 8.3\% across the three settings, respectively. This pattern suggests that the pretrained VLA prior provides useful generalization under milder distribution shifts but becomes insufficient when both the target object and referring expression are out of distribution. Base-L exhibits a different trade-off. On Tabletop, its Setting~III success improves substantially over Base (29.2\% vs.\ 0.0\%), but this comes with large drops in Settings~I and~II (37.5\% and 8.3\%, respectively). On Mobile Base, however, this advantage does not persist, with only 8.3\% success in Setting~III and an overall average of 5.5\%. If supported by our trajectory inspection, this behavior is consistent with the observation that the apparent gain comes partly from biased motion trajectories that occasionally reach the correct target rather than from reliable language-conditioned grounding.

% likely because the fine-tuning dataset is small and the augmented prompts introduce additional linguistic variation without sufficient trajectory diversity. When many prompts share similar syntactic structure but refer to different targets, the policy may learn a weaker association between the language condition and the corresponding action chunk. Although \textbf{Base}$^*$ performs better than \textbf{Base} on object-language-level out-of-domain tasks, rollout inspection suggests that these gains partly come from a biased default trajectory that occasionally reaches the correct target, rather than from robust language-conditioned grounding.

% \wayne{during the analysis, refer some numbers in the table, Table 2 is very big, more numbers could be referred, otherwise is a waste of space; for example, column-wise: different setting 1,2,3 comparison, Tabletop VS Mobile Base; row-wise: comparison with baselines.}
% As a result, errors in the planner output can directly propagate to the low-level controller.
Among the dual-system baselines, MOKA remains weak across both embodiments, achieving only 18.0\% average success on Tabletop and 8.3\% on Mobile Base. We observe that its failures often originate from inaccurate visual prompts produced by System~2, highlighting the difficulty of generating dense spatial guidance for the entire execution trajectory from a high-level planner. This challenge is further amplified by our forward-facing camera configuration, which covers a large workspace while also observing the robot itself, increasing perspective variation and visual clutter.
VP-VLA is the strongest baseline on Mobile Base, reaching 44.4\% average success and matching or exceeding Base in several settings. In particular, it improves substantially over Base in Mobile Setting~I (58.3\% vs.\ 33.3\%) and Setting~II (66.7\% vs.\ 50.0\%), but remains weak in Setting~III, with only 8.3\% success. A similar limitation appears on Tabletop, where VP-VLA obtains 4.2\% in Setting~III despite stronger performance in Settings~I and~II. These results suggest that auxiliary grounding supervision helps under moderate shifts but does not reliably align the generated action trajectory with the visual prompt under the strongest distribution shift.

In contrast, our method achieves the best or tied-best performance in every setting. The improvement is particularly pronounced in Setting~III, where it reaches 66.7\% on Tabletop compared with 29.2\% for the strongest baseline, and 41.7\% on Mobile Base compared with 8.3\% for the strongest baselines. Our method also maintains strong performance in Settings~I and~II, achieving 83.3\% and 66.7\% on both embodiments. We attribute these gains to two design choices: System~2 provides only sparse spatial cues, which are easier to obtain reliably than dense trajectory-level guidance, and the learned adapter uses these cues to steer the frozen policy output while preserving the pretrained VLA prior and fine-tuned manipulation skill.

% When transferring the policy from the table-top setup to the mobile base, we observe mixed effects. 

\begin{figure}[t]
    \vspace{4pt}
    \centering
    \includegraphics[width=1.0\linewidth, trim=50 195 60 110, clip]{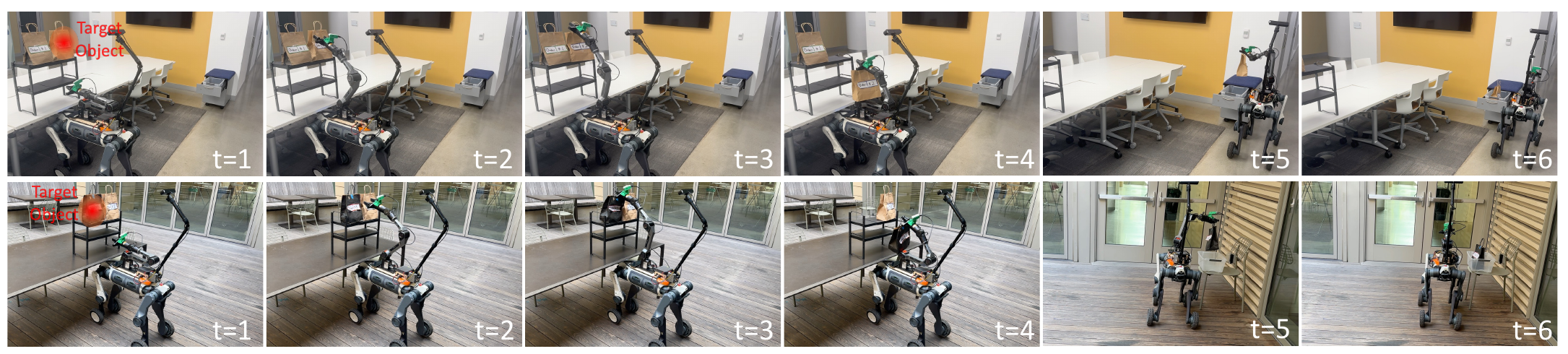}
    \vspace{-14pt}
    \caption{\textbf{Extension to a complete delivery pipeline.} We show example rollouts that combine manual navigation with autonomous manipulation. The red heatmap indicates the correctly grounded target object. After the mobile base is manually navigated to the workspace, our policy performs autonomous cue-conditioned manipulation to complete the task.}
    \label{fig:mobi_demo}
    \vspace{-12pt}
\end{figure}

\textbf{Qualitative results.} Figure~\ref{fig:comparison} compares representative rollouts under the unseen-object with novel-language setting on the mobile base, where successful execution requires both identifying the intended target and maintaining consistent control throughout manipulation. The Base policy exhibits weak target specificity, producing largely untargeted trajectories that often alternate between or reach toward both bags on the shelf. This behavior suggests that the pretrained policy retains a generic manipulation prior but lacks sufficient grounding to resolve the novel object--language correspondence. VP-VLA exhibits stronger sensitivity to the visual prompt and more frequently moves toward the relevant region. However, the prompted target is not consistently translated into a stable action trajectory, resulting in oscillatory motion, corrections toward competing objects, or eventual interaction with the wrong target. These rollouts suggest that injecting visual prompts into the policy improves spatial conditioning but does not by itself ensure persistent cue--action alignment under the stronger distribution shift of Setting~III. 
In contrast, our method maintains a consistent correspondence between the spatial cue and the generated action throughout execution. The adapter steers the frozen base policy toward the specified object while preserving the underlying manipulation behavior, yielding more direct approach trajectories and successful completion of the intended interaction.
Figure~\ref{fig:mobi_demo} further demonstrates that the cue-conditioned manipulation skill transfers across diverse scenes. Despite changes in scene layout and visual appearance during execution, the policy continues to follow the provided spatial cue and complete the corresponding manipulation.

% \textbf{Failure cases.} We also show failure cases in Figure~\ref{}. The results show that the failures mostly arise from 

\subsection{Analysis}
\begin{figure}[t]
\centering
\begin{minipage}[b]{0.4\linewidth}
    \centering
    \includegraphics[width=\linewidth, trim=0 0 0 5, clip]{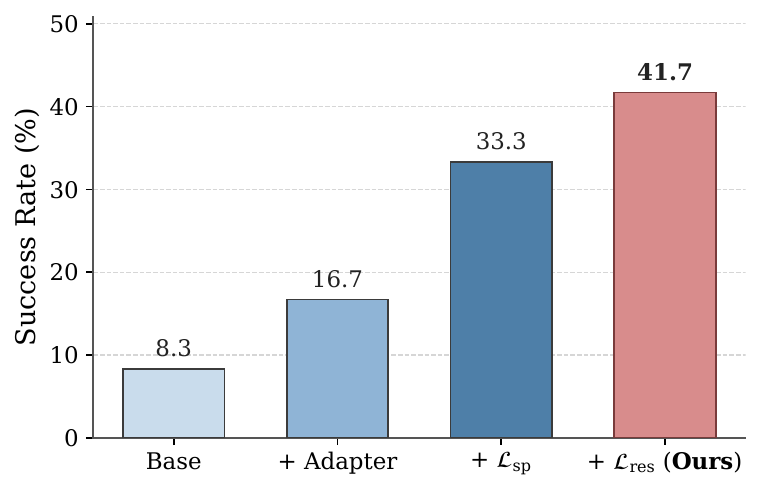}
    \vspace{-20pt}
    \captionof{figure}{\textbf{Effectiveness of components.}}
    \label{fig:component}
\end{minipage}
\hfill
\begin{minipage}[b]{0.59\linewidth}
    \centering
    \includegraphics[width=\linewidth, trim=240 150 245 105, clip]{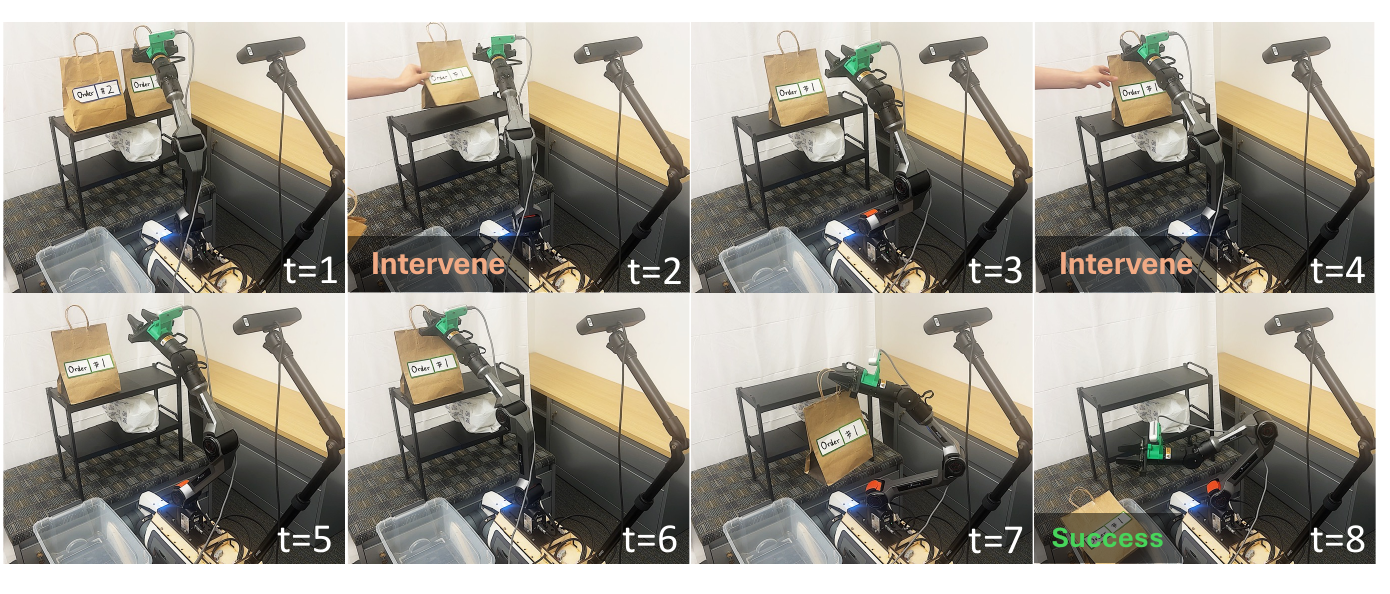}
    \vspace{-20pt}
    \captionof{figure}{\textbf{Reactivity to dynamically moving objects.}}
    \label{fig:dynamic}
\end{minipage}
\vspace{-24pt}
\end{figure}

\textbf{Component effectiveness.} Figure~\ref{fig:component} evaluates the contribution of each learning component under Setting~III. Starting from the Base policy at 8.3\% success, introducing the adapter with spatial cue conditioning increases performance to 16.7\%, showing that cue-conditioned residual adaptation alone provides only limited improvement. Adding the spatial contrastive loss $\mathcal{L}_{\mathrm{sp}}$ further raises success to 33.3\%, a 16.6 percentage-point gain, indicating that explicitly learning spatially discriminative cue representations is critical for effective steering. Incorporating the residual contrastive loss $\mathcal{L}_{\mathrm{res}}$ yields an additional improvement to 41.7\%, producing the best overall result. Together, these results show that the adapter benefits substantially from both contrastive objectives, with $\mathcal{L}_{\mathrm{sp}}$ providing the largest individual gain and $\mathcal{L}_{\mathrm{res}}$ further refining the learned steering signal.

% noContrast retains the entire G adapter architecture (cue encoder $f_{\text{cue}}$ + output-FiLM head $W_{\text{vfilm}}$) and trains both via the flow loss; only the contrastive auxiliary losses (CAB, RAB) are switched off. So it differs from the base policy in three ways: (1) it has trainable  cue-pathway parameters that the base lacks, (2) it applies a multiplicative cue modulation $\vused = \vbase \odot (1 + \gamma_v) + \beta_v$ at deploy  whereas the base emits $\vbase$ directly, and (3) the cue is informationally present at the velocity head — just shaped only by flow loss on swap-triggered batches, not by spatially-discriminative contrastive supervision. We predict the trained cue effect is non-zero but weaker than G_v2's, and the comparison G_v2 vs noContrast quantifies how much of the cue-following ability is owed to contrastive shaping vs. flow-loss alone.

\textbf{Effect of spatial cue format.}
We further study the effect of different spatial cue formats. In addition to the soft Gaussian heatmap, we evaluate a binary target mask, defined as a segmentation silhouette at patch resolution, and an axis-aligned bounding box, which captures coarse object extent. The Gaussian heatmap and binary mask yield similar behavior and performance, indicating that localized target cues are sufficient for effective steering. In contrast, the bounding box performs worse, likely because its coarse boundaries can direct the robot toward invalid or non-manipulable object regions. This suggests that fine-grained spatial cues are more effective than the coarse box-level one.
% The three cue representations in this ablation differ in the granularity at which they localize the target in the patch grid, while sharing the same underlying SAM3 perception output. cueGaussian (the default) is a soft heatmap centered at the object's centroid, providing peak-localized but smoothly-extended spatial mass. cueBinary is the object's segmentation silhouette at patch resolution, preserving the exact shape of the target with sharp boundaries. cueBBox is the object's axis-aligned bounding region with sharp boundaries but no shape information beyond coarse extent. Together they form a controlled spectrum from soft-centroid to exact-shape to bbox-only localization, isolating which property of the spatial cue—smoothness, shape fidelity, or region-level coarseness—the adapter actually exploits.

% \textbf{Counterfactual prompting.} Prompt with a different focus than the language -> We cannot do this... our prompt is fixed

\textbf{Reactivity to dynamically changing environments.}
Figure~\ref{fig:dynamic} demonstrates the benefit of using a large-scale VLA as the System~1 controller by perturbing the target object's position during execution. With the spatial cue grounded only once at the beginning of the rollout, our method still reacts to environmental changes without re-querying the high-level planner. This reactivity is inherited from the underlying VLA and preserved by our adapter, highlighting an advantage over conventional geometric controllers that rely on fixed trajectories or repeated replanning.
% In domain benefit of the VLA seeing fixed language input even when the user prompt changes dramastically

\textbf{Simulation evaluation.} To complement the real-world experiments, we further evaluate our method in simulation using \textbf{LIBERO-PRO}~\citep{libero_pro}. We adopt the Object-suite task-generalization setting, which evaluates novel language--object mappings and aligns with our objective of improving generalization to unseen task compositions. Across 200 evaluation episodes, our method achieves the highest success rate at 15.5\%, compared with 10.5\% for Base, 9.5\% for Base-L, and 10.5\% for VP-VLA. This controlled benchmark also allows us to examine the effect of the regularization strength $\lambda_{\mathrm{reg}}$. For $\lambda_{\mathrm{reg}}\in \{0,0.01,0.03,0.1\}$, the corresponding success rates are 11.5\%, 15.5\%, 12.5\%, and 10.5\%, respectively. The best performance at $\lambda_{\mathrm{reg}}=0.01$ reflects the intended trade-off: without regularization, steering can overcorrect the base policy, whereas overly strong regularization suppresses useful steering and reduces its benefit.

\section{Conclusion}
We introduced a spatial-cue-steered dual-system framework for open-world delivery manipulation. By replacing the geometric low-level controller with a pretrained flow-matching VLA, our framework improves robustness to visual perturbations and reactivity to dynamic environments. It converts System~2 outputs into action-steering signals via a lightweight adapter while keeping the VLA frozen, preserving its manipulation prior and improving generalization to novel language--object mappings. Across tabletop and mobile manipulation evaluations, our method improves in-domain and out-of-domain performance with negligible additional inference cost, showing that spatial-cue steering provides a practical interface between open-vocabulary grounding and reactive VLA control.

\section{Limitations}
Our method depends on the underlying flow-matching policy and remains sensitive to out-of-distribution affordance regions and object positions, constraining the mobile base's stopping location and orientation relative to the target. Future work will address more complex delivery scenarios, including human interaction and handover, item selection from densely clustered bags, and tighter integration with navigation policies to improve manipulation reliability.

% Performance may degrade when the mobile base stops at a pose from which the target is outside the policy's reliable manipulation range.
% is not without limitations. First, the affordance area our method can handle is constrained. With objects that has significantly different affordance than the human demonstrations used for finetuning, our method cannot handle such cases. Fortunately, the objects within the delivery task are diverse but similar in general category (bags with affordance at the top). Second, the spatial coverage of the spatial cue steering is limited by the base policy, where we cannot extend the robot to manipulate on objects that are spatially out-of-distribution. However, this can be alleviated by extending the amount of demonstration data to include action trajectories that cover the whole distribution of the robot's valid workspace. Future work would involve aspects such as interaction with human and delivering bags to human hands, finding the right bag within a clustered amount of bags, including more manipulations skills such as loading the takeout onto the robot, and integration with an underlying navigation policy. 
% \wayne{describe a more general delivery scenario and say this is the future work: hand-to-hand delivery, no reliability to \# tags, including load and unload, navigation.}

%===============================================================================

\clearpage
% The acknowledgments are automatically included only in the final and preprint versions of the paper.
\acknowledgments{This project was supported in part by NSF grants IIS-2339769, IIS-2525840, CNS-2432534, and ECCS-2514574. We acknowledge the use of the DEEP Robotics LYNX M20 wheeled quadruped robot in the experiments. The views and conclusions expressed in this article are solely those of the authors and do not necessarily reflect those of the funding agencies or supporting organizations.
}
%If a paper is accepted, the final camera-ready version will (and probably should) include acknowledgments. All acknowledgments go at the end of the paper, including thanks to reviewers who gave useful comments, to colleagues who contributed to the ideas, and to funding agencies and corporate sponsors that provided financial support.}

%===============================================================================

\bibliography{example}  % .bib

@misc{tracevla,
      title={TraceVLA: Visual Trace Prompting Enhances Spatial-Temporal Awareness for Generalist Robotic Policies}, 
      author={Ruijie Zheng and Yongyuan Liang and Shuaiyi Huang and Jianfeng Gao and Hal Daumé III and Andrey Kolobov and Furong Huang and Jianwei Yang},
      year={2025},
      eprint={2412.10345},
      archivePrefix={arXiv},
      primaryClass={cs.RO},
      url={https://arxiv.org/abs/2412.10345}, 
}

@misc{vp-vla,
      title={VP-VLA: Visual Prompting as an Interface for Vision-Language-Action Models}, 
      author={Zixuan Wang and Yuxin Chen and Yuqi Liu and Jinhui Ye and Pengguang Chen and Changsheng Lu and Shu Liu and Bei Yu and Jiaya Jia},
      year={2026},
      eprint={2603.22003},
      archivePrefix={arXiv},
      primaryClass={cs.RO},
      url={https://arxiv.org/abs/2603.22003}, 
}

@misc{moka,
      title={MOKA: Open-World Robotic Manipulation through Mark-Based Visual Prompting}, 
      author={Fangchen Liu and Kuan Fang and Pieter Abbeel and Sergey Levine},
      year={2024},
      eprint={2403.03174},
      archivePrefix={arXiv},
      primaryClass={cs.RO},
      url={https://arxiv.org/abs/2403.03174}, 
}

@misc{robopoint,
      title={RoboPoint: A Vision-Language Model for Spatial Affordance Prediction for Robotics}, 
      author={Wentao Yuan and Jiafei Duan and Valts Blukis and Wilbert Pumacay and Ranjay Krishna and Adithyavairavan Murali and Arsalan Mousavian and Dieter Fox},
      year={2024},
      eprint={2406.10721},
      archivePrefix={arXiv},
      primaryClass={cs.RO},
      url={https://arxiv.org/abs/2406.10721}, 
}

@misc{rekep,
      title={ReKep: Spatio-Temporal Reasoning of Relational Keypoint Constraints for Robotic Manipulation}, 
      author={Wenlong Huang and Chen Wang and Yunzhu Li and Ruohan Zhang and Li Fei-Fei},
      year={2024},
      eprint={2409.01652},
      archivePrefix={arXiv},
      primaryClass={cs.RO},
      url={https://arxiv.org/abs/2409.01652}, 
}

@misc{crayonrobo,
      title={CrayonRobo: Object-Centric Prompt-Driven Vision-Language-Action Model for Robotic Manipulation}, 
      author={Xiaoqi Li and Lingyun Xu and Mingxu Zhang and Jiaming Liu and Yan Shen and Iaroslav Ponomarenko and Jiahui Xu and Liang Heng and Siyuan Huang and Shanghang Zhang and Hao Dong},
      year={2025},
      eprint={2505.02166},
      archivePrefix={arXiv},
      primaryClass={cs.RO},
      url={https://arxiv.org/abs/2505.02166}, 
}

@misc{vap,
      title={Bring My Cup! Personalizing Vision-Language-Action Models with Visual Attentive Prompting}, 
      author={Sangoh Lee and Sangwoo Mo and Wook-Shin Han},
      year={2026},
      eprint={2512.20014},
      archivePrefix={arXiv},
      primaryClass={cs.RO},
      url={https://arxiv.org/abs/2512.20014}, 
}

@misc{steer_rl,
      title={Steering Your Generalists: Improving Robotic Foundation Models via Value Guidance}, 
      author={Mitsuhiko Nakamoto and Oier Mees and Aviral Kumar and Sergey Levine},
      year={2025},
      eprint={2410.13816},
      archivePrefix={arXiv},
      primaryClass={cs.RO},
      url={https://arxiv.org/abs/2410.13816}, 
}

@misc{lpb,
      title={Latent Policy Barrier: Learning Robust Visuomotor Policies by Staying In-Distribution}, 
      author={Zhanyi Sun and Shuran Song},
      year={2025},
      eprint={2508.05941},
      archivePrefix={arXiv},
      primaryClass={cs.RO},
      url={https://arxiv.org/abs/2508.05941}, 
}

@misc{dynaguide,
      title={DynaGuide: Steering Diffusion Polices with Active Dynamic Guidance}, 
      author={Maximilian Du and Shuran Song},
      year={2025},
      eprint={2506.13922},
      archivePrefix={arXiv},
      primaryClass={cs.RO},
      url={https://arxiv.org/abs/2506.13922}, 
}

@misc{human_loop_1,
      title={Predictive Preference Learning from Human Interventions}, 
      author={Haoyuan Cai and Zhenghao Peng and Bolei Zhou},
      year={2025},
      eprint={2510.01545},
      archivePrefix={arXiv},
      primaryClass={cs.LG},
      url={https://arxiv.org/abs/2510.01545}, 
}

@misc{human_loop_2,
      title={From Foresight to Forethought: VLM-In-the-Loop Policy Steering via Latent Alignment}, 
      author={Yilin Wu and Ran Tian and Gokul Swamy and Andrea Bajcsy},
      year={2025},
      eprint={2502.01828},
      archivePrefix={arXiv},
      primaryClass={cs.RO},
      url={https://arxiv.org/abs/2502.01828}, 
}

@misc{human_loop_3,
      title={Learning from Active Human Involvement through Proxy Value Propagation}, 
      author={Zhenghao Peng and Wenjie Mo and Chenda Duan and Quanyi Li and Bolei Zhou},
      year={2025},
      eprint={2502.03369},
      archivePrefix={arXiv},
      primaryClass={cs.AI},
      url={https://arxiv.org/abs/2502.03369}, 
}

@inproceedings{mpc_safety,
   title={Generalizing Safety Beyond Collision-Avoidance via Latent-Space Reachability Analysis},
   url={http://dx.doi.org/10.15607/RSS.2025.XXI.113},
   DOI={10.15607/rss.2025.xxi.113},
   booktitle={Robotics: Science and Systems XXI},
   publisher={Robotics: Science and Systems Foundation},
   author={Nakamura, Kensuke and Peters, Lasse and Bajcsy, Andrea},
   year={2025},
   month=jun, collection={RSS2025} 
}

@misc{omniguide,
      title={OmniGuide: Universal Guidance Fields for Enhancing Generalist Robot Policies}, 
      author={Yunzhou Song and Long Le and Yong-Hyun Park and Jie Wang and Junyao Shi and Lingjie Liu and Jiatao Gu and Eric Eaton and Dinesh Jayaraman and Kostas Daniilidis},
      year={2026},
      eprint={2603.10052},
      archivePrefix={arXiv},
      primaryClass={cs.RO},
      url={https://arxiv.org/abs/2603.10052}, 
}

@misc{dynamics_model,
      title={Learning Latent Dynamics for Planning from Pixels}, 
      author={Danijar Hafner and Timothy Lillicrap and Ian Fischer and Ruben Villegas and David Ha and Honglak Lee and James Davidson},
      year={2019},
      eprint={1811.04551},
      archivePrefix={arXiv},
      primaryClass={cs.LG},
      url={https://arxiv.org/abs/1811.04551}, 
}

@misc{daydreamer,
      title={DayDreamer: World Models for Physical Robot Learning}, 
      author={Philipp Wu and Alejandro Escontrela and Danijar Hafner and Ken Goldberg and Pieter Abbeel},
      year={2022},
      eprint={2206.14176},
      archivePrefix={arXiv},
      primaryClass={cs.RO},
      url={https://arxiv.org/abs/2206.14176}, 
}

@misc{pi05,
      title={$\pi_{0.5}$: a Vision-Language-Action Model with Open-World Generalization}, 
      author={Physical Intelligence and Kevin Black and Noah Brown and James Darpinian and Karan Dhabalia and Danny Driess and Adnan Esmail and Michael Equi and Chelsea Finn and Niccolo Fusai and Manuel Y. Galliker and Dibya Ghosh and Lachy Groom and Karol Hausman and Brian Ichter and Szymon Jakubczak and Tim Jones and Liyiming Ke and Devin LeBlanc and Sergey Levine and Adrian Li-Bell and Mohith Mothukuri and Suraj Nair and Karl Pertsch and Allen Z. Ren and Lucy Xiaoyang Shi and Laura Smith and Jost Tobias Springenberg and Kyle Stachowicz and James Tanner and Quan Vuong and Homer Walke and Anna Walling and Haohuan Wang and Lili Yu and Ury Zhilinsky},
      year={2025},
      eprint={2504.16054},
      archivePrefix={arXiv},
      primaryClass={cs.LG},
      url={https://arxiv.org/abs/2504.16054}, 
}

@misc{qwen3vl,
      title={Qwen3-VL Technical Report}, 
      author={Shuai Bai and Yuxuan Cai and Ruizhe Chen and Keqin Chen and Xionghui Chen and Zesen Cheng and Lianghao Deng and Wei Ding and Chang Gao and Chunjiang Ge and Wenbin Ge and Zhifang Guo and Qidong Huang and Jie Huang and Fei Huang and Binyuan Hui and Shutong Jiang and Zhaohai Li and Mingsheng Li and Mei Li and Kaixin Li and Zicheng Lin and Junyang Lin and Xuejing Liu and Jiawei Liu and Chenglong Liu and Yang Liu and Dayiheng Liu and Shixuan Liu and Dunjie Lu and Ruilin Luo and Chenxu Lv and Rui Men and Lingchen Meng and Xuancheng Ren and Xingzhang Ren and Sibo Song and Yuchong Sun and Jun Tang and Jianhong Tu and Jianqiang Wan and Peng Wang and Pengfei Wang and Qiuyue Wang and Yuxuan Wang and Tianbao Xie and Yiheng Xu and Haiyang Xu and Jin Xu and Zhibo Yang and Mingkun Yang and Jianxin Yang and An Yang and Bowen Yu and Fei Zhang and Hang Zhang and Xi Zhang and Bo Zheng and Humen Zhong and Jingren Zhou and Fan Zhou and Jing Zhou and Yuanzhi Zhu and Ke Zhu},
      year={2025},
      eprint={2511.21631},
      archivePrefix={arXiv},
      primaryClass={cs.CV},
      url={https://arxiv.org/abs/2511.21631}, 
}

@misc{sam3,
      title={SAM 3: Segment Anything with Concepts}, 
      author={Nicolas Carion and Laura Gustafson and Yuan-Ting Hu and Shoubhik Debnath and Ronghang Hu and Didac Suris and Chaitanya Ryali and Kalyan Vasudev Alwala and Haitham Khedr and Andrew Huang and Jie Lei and Tengyu Ma and Baishan Guo and Arpit Kalla and Markus Marks and Joseph Greer and Meng Wang and Peize Sun and Roman Rädle and Triantafyllos Afouras and Effrosyni Mavroudi and Katherine Xu and Tsung-Han Wu and Yu Zhou and Liliane Momeni and Rishi Hazra and Shuangrui Ding and Sagar Vaze and Francois Porcher and Feng Li and Siyuan Li and Aishwarya Kamath and Ho Kei Cheng and Piotr Dollár and Nikhila Ravi and Kate Saenko and Pengchuan Zhang and Christoph Feichtenhofer},
      year={2026},
      eprint={2511.16719},
      archivePrefix={arXiv},
      primaryClass={cs.CV},
      url={https://arxiv.org/abs/2511.16719}, 
}

@misc{openvla,
      title={OpenVLA: An Open-Source Vision-Language-Action Model}, 
      author={Moo Jin Kim and Karl Pertsch and Siddharth Karamcheti and Ted Xiao and Ashwin Balakrishna and Suraj Nair and Rafael Rafailov and Ethan Foster and Grace Lam and Pannag Sanketi and Quan Vuong and Thomas Kollar and Benjamin Burchfiel and Russ Tedrake and Dorsa Sadigh and Sergey Levine and Percy Liang and Chelsea Finn},
      year={2024},
      eprint={2406.09246},
      archivePrefix={arXiv},
      primaryClass={cs.RO},
      url={https://arxiv.org/abs/2406.09246}, 
}

@misc{rt2,
      title={RT-2: Vision-Language-Action Models Transfer Web Knowledge to Robotic Control}, 
      author={Anthony Brohan and Noah Brown and Justice Carbajal and Yevgen Chebotar and Xi Chen and Krzysztof Choromanski and Tianli Ding and Danny Driess and Avinava Dubey and Chelsea Finn and Pete Florence and Chuyuan Fu and Montse Gonzalez Arenas and Keerthana Gopalakrishnan and Kehang Han and Karol Hausman and Alexander Herzog and Jasmine Hsu and Brian Ichter and Alex Irpan and Nikhil Joshi and Ryan Julian and Dmitry Kalashnikov and Yuheng Kuang and Isabel Leal and Lisa Lee and Tsang-Wei Edward Lee and Sergey Levine and Yao Lu and Henryk Michalewski and Igor Mordatch and Karl Pertsch and Kanishka Rao and Krista Reymann and Michael Ryoo and Grecia Salazar and Pannag Sanketi and Pierre Sermanet and Jaspiar Singh and Anikait Singh and Radu Soricut and Huong Tran and Vincent Vanhoucke and Quan Vuong and Ayzaan Wahid and Stefan Welker and Paul Wohlhart and Jialin Wu and Fei Xia and Ted Xiao and Peng Xu and Sichun Xu and Tianhe Yu and Brianna Zitkovich},
      year={2023},
      eprint={2307.15818},
      archivePrefix={arXiv},
      primaryClass={cs.RO},
      url={https://arxiv.org/abs/2307.15818}, 
}

@misc{ado_noising,
      title={Much Ado About Noising: Dispelling the Myths of Generative Robotic Control}, 
      author={Chaoyi Pan and Giri Anantharaman and Nai-Chieh Huang and Claire Jin and Daniel Pfrommer and Chenyang Yuan and Frank Permenter and Guannan Qu and Nicholas Boffi and Guanya Shi and Max Simchowitz},
      year={2026},
      eprint={2512.01809},
      archivePrefix={arXiv},
      primaryClass={cs.RO},
      url={https://arxiv.org/abs/2512.01809}, 
}

@misc{gr00t,
      title={GR00T N1: An Open Foundation Model for Generalist Humanoid Robots}, 
      author={NVIDIA and : and Johan Bjorck and Fernando Castañeda and Nikita Cherniadev and Xingye Da and Runyu Ding and Linxi "Jim" Fan and Yu Fang and Dieter Fox and Fengyuan Hu and Spencer Huang and Joel Jang and Zhenyu Jiang and Jan Kautz and Kaushil Kundalia and Lawrence Lao and Zhiqi Li and Zongyu Lin and Kevin Lin and Guilin Liu and Edith Llontop and Loic Magne and Ajay Mandlekar and Avnish Narayan and Soroush Nasiriany and Scott Reed and You Liang Tan and Guanzhi Wang and Zu Wang and Jing Wang and Qi Wang and Jiannan Xiang and Yuqi Xie and Yinzhen Xu and Zhenjia Xu and Seonghyeon Ye and Zhiding Yu and Ao Zhang and Hao Zhang and Yizhou Zhao and Ruijie Zheng and Yuke Zhu},
      year={2025},
      eprint={2503.14734},
      archivePrefix={arXiv},
      primaryClass={cs.RO},
      url={https://arxiv.org/abs/2503.14734}, 
}

@misc{smolvla,
      title={SmolVLA: A Vision-Language-Action Model for Affordable and Efficient Robotics}, 
      author={Mustafa Shukor and Dana Aubakirova and Francesco Capuano and Pepijn Kooijmans and Steven Palma and Adil Zouitine and Michel Aractingi and Caroline Pascal and Martino Russi and Andres Marafioti and Simon Alibert and Matthieu Cord and Thomas Wolf and Remi Cadene},
      year={2025},
      eprint={2506.01844},
      archivePrefix={arXiv},
      primaryClass={cs.LG},
      url={https://arxiv.org/abs/2506.01844}, 
}

@article{sam2,
  title={SAM 2: Segment Anything in Images and Videos},
  author={Ravi, Nikhila and Gabeur, Valentin and Hu, Yuan-Ting and Hu, Ronghang and Ryali, Chaitanya and Ma, Tengyu and Khedr, Haitham and R{\"a}dle, Roman and Rolland, Chloe and Gustafson, Laura and Mintun, Eric and Pan, Junting and Alwala, Kalyan Vasudev and Carion, Nicolas and Wu, Chao-Yuan and Girshick, Ross and Doll{\'a}r, Piotr and Feichtenhofer, Christoph},
  journal={arXiv preprint arXiv:2408.00714},
  url={https://arxiv.org/abs/2408.00714},
  year={2024}
}

@article{vp_1,
  title={Visual prompting via image inpainting},
  author={Bar, Amir and Gandelsman, Yossi and Darrell, Trevor and Globerson, Amir and Efros, Alexei},
  journal={Advances in neural information processing systems},
  volume={35},
  pages={25005--25017},
  year={2022}
}

@misc{vp_2,
      title={Set-of-Mark Prompting Unleashes Extraordinary Visual Grounding in GPT-4V}, 
      author={Jianwei Yang and Hao Zhang and Feng Li and Xueyan Zou and Chunyuan Li and Jianfeng Gao},
      year={2023},
      eprint={2310.11441},
      archivePrefix={arXiv},
      primaryClass={cs.CV},
      url={https://arxiv.org/abs/2310.11441}, 
}

@misc{internvla,
      title={InternVLA-M1: A Spatially Guided Vision-Language-Action Framework for Generalist Robot Policy}, 
      author={Xinyi Chen and Yilun Chen and Yanwei Fu and Ning Gao and Jiaya Jia and Weiyang Jin and Hao Li and Yao Mu and Jiangmiao Pang and Yu Qiao and Yang Tian and Bin Wang and Bolun Wang and Fangjing Wang and Hanqing Wang and Tai Wang and Ziqin Wang and Xueyuan Wei and Chao Wu and Shuai Yang and Jinhui Ye and Junqiu Yu and Jia Zeng and Jingjing Zhang and Jinyu Zhang and Shi Zhang and Feng Zheng and Bowen Zhou and Yangkun Zhu},
      year={2025},
      eprint={2510.13778},
      archivePrefix={arXiv},
      primaryClass={cs.RO},
      url={https://arxiv.org/abs/2510.13778}, 
}

@misc{pi0,
      title={$\pi_0$: A Vision-Language-Action Flow Model for General Robot Control}, 
      author={Kevin Black and Noah Brown and Danny Driess and Adnan Esmail and Michael Equi and Chelsea Finn and Niccolo Fusai and Lachy Groom and Karol Hausman and Brian Ichter and Szymon Jakubczak and Tim Jones and Liyiming Ke and Sergey Levine and Adrian Li-Bell and Mohith Mothukuri and Suraj Nair and Karl Pertsch and Lucy Xiaoyang Shi and James Tanner and Quan Vuong and Anna Walling and Haohuan Wang and Ury Zhilinsky},
      year={2026},
      eprint={2410.24164},
      archivePrefix={arXiv},
      primaryClass={cs.LG},
      url={https://arxiv.org/abs/2410.24164}, 
}

@misc{hirobot,
      title={Hi Robot: Open-Ended Instruction Following with Hierarchical Vision-Language-Action Models}, 
      author={Lucy Xiaoyang Shi and Brian Ichter and Michael Equi and Liyiming Ke and Karl Pertsch and Quan Vuong and James Tanner and Anna Walling and Haohuan Wang and Niccolo Fusai and Adrian Li-Bell and Danny Driess and Lachy Groom and Sergey Levine and Chelsea Finn},
      year={2025},
      eprint={2502.19417},
      archivePrefix={arXiv},
      primaryClass={cs.RO},
      url={https://arxiv.org/abs/2502.19417}, 
}

@misc{pointvla,
      title={PointVLA: Injecting the 3D World into Vision-Language-Action Models}, 
      author={Chengmeng Li and Junjie Wen and Yan Peng and Yaxin Peng and Feifei Feng and Yichen Zhu},
      year={2025},
      eprint={2503.07511},
      archivePrefix={arXiv},
      primaryClass={cs.RO},
      url={https://arxiv.org/abs/2503.07511}, 
}

@misc{spatialvla,
      title={SpatialVLA: Exploring Spatial Representations for Visual-Language-Action Model}, 
      author={Delin Qu and Haoming Song and Qizhi Chen and Yuanqi Yao and Xinyi Ye and Yan Ding and Zhigang Wang and JiaYuan Gu and Bin Zhao and Dong Wang and Xuelong Li},
      year={2025},
      eprint={2501.15830},
      archivePrefix={arXiv},
      primaryClass={cs.RO},
      url={https://arxiv.org/abs/2501.15830}, 
}

@misc{vipa,
      title={Spatial-Aware VLA Pretraining through Visual-Physical Alignment from Human Videos}, 
      author={Yicheng Feng and Wanpeng Zhang and Ye Wang and Hao Luo and Haoqi Yuan and Sipeng Zheng and Zongqing Lu},
      year={2025},
      eprint={2512.13080},
      archivePrefix={arXiv},
      primaryClass={cs.RO},
      url={https://arxiv.org/abs/2512.13080}, 
}

@misc{vca,
      title={VCA: Vision-Click-Action Framework for Precise Manipulation of Segmented Objects in Target Ambiguous Environments}, 
      author={Donggeon Kim and Seungwon Jan and Hyeonjun Park and Daegyu Lim},
      year={2026},
      eprint={2602.23583},
      archivePrefix={arXiv},
      primaryClass={cs.RO},
      url={https://arxiv.org/abs/2602.23583}, 
}

@misc{lora,
      title={LoRA: Low-Rank Adaptation of Large Language Models}, 
      author={Edward J. Hu and Yelong Shen and Phillip Wallis and Zeyuan Allen-Zhu and Yuanzhi Li and Shean Wang and Lu Wang and Weizhu Chen},
      year={2021},
      eprint={2106.09685},
      archivePrefix={arXiv},
      primaryClass={cs.CL},
      url={https://arxiv.org/abs/2106.09685}, 
}

@misc{colosseum,
      title={THE COLOSSEUM: A Benchmark for Evaluating Generalization for Robotic Manipulation}, 
      author={Wilbert Pumacay and Ishika Singh and Jiafei Duan and Ranjay Krishna and Jesse Thomason and Dieter Fox},
      year={2024},
      eprint={2402.08191},
      archivePrefix={arXiv},
      primaryClass={cs.RO},
      url={https://arxiv.org/abs/2402.08191}, 
}

@misc{kpam,
      title={kPAM: KeyPoint Affordances for Category-Level Robotic Manipulation}, 
      author={Lucas Manuelli and Wei Gao and Peter Florence and Russ Tedrake},
      year={2019},
      eprint={1903.06684},
      archivePrefix={arXiv},
      primaryClass={cs.RO},
      url={https://arxiv.org/abs/1903.06684}, 
}

@misc{action_language,
      title={Actions as Language: Fine-Tuning VLMs into VLAs Without Catastrophic Forgetting}, 
      author={Asher J. Hancock and Xindi Wu and Lihan Zha and Olga Russakovsky and Anirudha Majumdar},
      year={2025},
      eprint={2509.22195},
      archivePrefix={arXiv},
      primaryClass={cs.RO},
      url={https://arxiv.org/abs/2509.22195}, 
}

@misc{libero_pro,
      title={LIBERO-PRO: Towards Robust and Fair Evaluation of Vision-Language-Action Models Beyond Memorization}, 
      author={Xueyang Zhou and Yangming Xu and Guiyao Tie and Yongchao Chen and Guowen Zhang and Duanfeng Chu and Pan Zhou and Lichao Sun},
      year={2026},
      eprint={2510.03827},
      archivePrefix={arXiv},
      primaryClass={cs.CV},
      url={https://arxiv.org/abs/2510.03827}, 
}

\newpage
\appendix
\section{Method Details}
\subsection{MOKA}
MOKA was originally designed for tabletop manipulation with top-down observations and 2D planar affordance prediction, which does not directly match our third-person shelf-picking setting with deformable bag objects and non-planar interactions. Rather than fully re-engineering MOKA for this domain, we evaluate its high-level planning capability under an oracle execution assumption: a prediction is considered successful if the generated 2D affordance points correspond to semantically correct and physically feasible interaction locations.

For consistency across methods, we standardize the perception stack with Qwen3-VL~\citep{qwen3vl} and SAM2.1~\citep{sam2}, and adapt the prompting strategy to our shelf-manipulation tasks. We also evaluate MOKA under its 3-shot in-context configuration, corresponding to the strongest setting reported in the original work. Despite these adaptations and the oracle execution assumption, performance remains limited due to target misclassification, incorrect grasp-point prediction, and infeasible proposed motions. These results indicate that MOKA's 2D affordance representation has limited transferability to our shelf-based delivery manipulation setting.

\subsection{VP-VLA}
The original VP-VLA method is designed for autoregressive VLAs and is not directly formulated for flow-matching policies. We adapt its visual-prompting mechanism by providing the generated visual prompt as an additional visual input to the VLA and training the policy with auxiliary grounding supervision. This baseline differs from our method primarily in how the System~2 visual signal is used: VP-VLA treats it as an input-level prompt, whereas our method converts the spatial cue into an action-steering signal that modulates the frozen policy output. For consistency across methods, we use the same perception stack with Qwen3-VL~\citep{qwen3vl} and SAM2.1~\citep{sam2}.

\subsection{Ours}
We rewrite each user instruction into a standardized command format compatible with the VLA's training distribution, removing task-irrelevant details while keeping the target specification. For demonstrations involving mobile navigation, we divide execution into three stages. First, the mobile base remains stationary while the robot arm performs autonomous manipulation. Once the gripper grasps the target object, policy inference is paused and a hard-coded arm motion moves the object to a predefined holding pose for stable transport. The grasping moment is detected when the gripper jaw width falls below a fixed threshold. Second, we manually teleoperate or use map-based autonomous navigation to guide the mobile base toward the target placement location. Third, policy inference is resumed to allow the robot to autonomously complete the placement stage. The demonstrations are recorded in both indoor and outdoor environments, with diverse backgrounds and lighting conditions. We observe that policy performance and steering effectiveness decrease under these out-of-domain conditions, but the model still shows a consistent steering trend and retains the ability to identify target objects and complete the tasks.

\section{Experiment Details}
\subsection{Hardware Setup}
\begin{figure}[h]
    \centering
    % \vspace{-6pt}
    \includegraphics[width=0.6\linewidth, trim=90 0 160 10, clip]{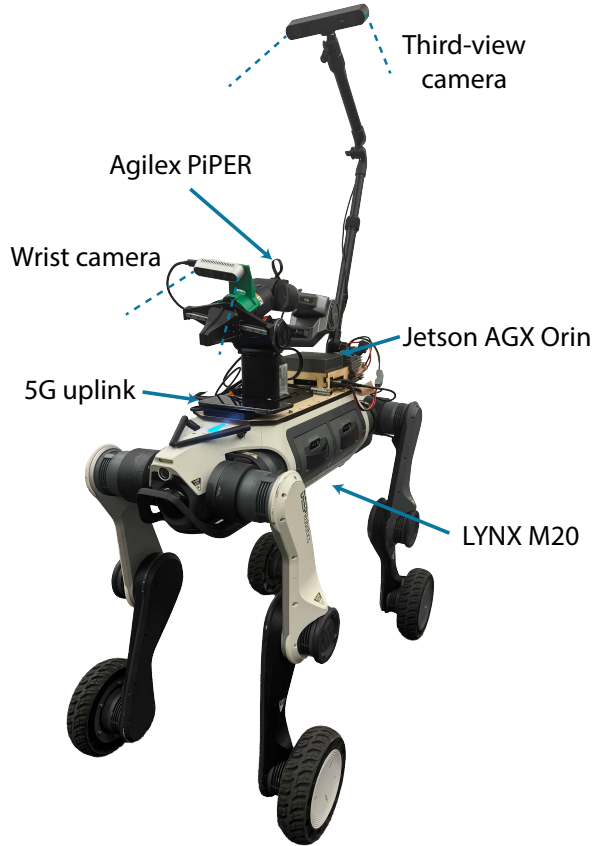}
    \vspace{-16pt}
    \caption{Robot hardware setup.}
    \label{fig:robot_hardware}
    \vspace{-8pt}
\end{figure}

Our experimental platform consists of a DEEP Robotics Lynx M20 Pro wheeled quadruped integrated with an AgileX Robotics PiPER robotic arm. We use a Stereolabs ZED 2 stereo camera for the third-person view and an Intel RealSense D435i depth camera as a wrist-mounted camera on the manipulator, although depth and stereo measurements are not used in this paper. A Jetson AGX Orin mounted on the robot coordinates communication between the robot hardware and a remote workstation running the policy on an RTX 5080 GPU. For outdoor experiments, network access is provided by USB tethering the Jetson to a mobile phone, which serves as a 5G uplink. The final hardware layout is shown in Figure~\ref{fig:robot_hardware}.

The Jetson AGX Orin serves as a lightweight communication bridge rather than an inference device. It captures both camera streams and transmits them to the remote workstation through local network communication. The workstation receives the video streams, runs the policy, and sends action commands back through the same network path to the Jetson, which forwards them to the robot arm. This setup incurs approximately 80ms round-trip latency, modestly higher than wired operation.
% Because both endpoints are behind NAT, packets are relayed through a Tailscale DERP server rather than a direct WireGuard tunnel. 

\subsection{Evaluation Setup}
To evaluate robustness to linguistic variation and compositional generalization, we construct a fixed set of prompt templates covering the three evaluation settings described in the main paper. Prompts are generated compositionally by varying grasp verbs, object descriptors, and placement phrasing while preserving the same underlying pick-and-place task semantics. Representative prompt templates are shown in Table~\ref{tab:prompt_list}.

\begin{table}[ht]
\centering
\small
\begin{tabular}{p{2.9cm} p{10.2cm}}
\toprule
\textbf{Evaluation Setting} & \textbf{Prompt Templates} \\
\midrule

Setting I \& II &
\begin{tabular}[t]{@{}l@{}}
Pick the \slot{descriptor} bag and put it into the box \\
Grab \slot{descriptor} bag and place it in the box \\
Take the \slot{descriptor} bag and drop it into the box \\
Move the \slot{descriptor} bag into the box \\
Put \slot{descriptor} bag into the box \\
Place the \slot{descriptor} bag in the box
\end{tabular}
\\

\midrule

Setting III &
\begin{tabular}[t]{@{}l@{}}
Put bag with order number \slot{value} into the box \\
Grab bag with number \slot{value} into the box \\
Pick order number \slot{value} bag and place it into the box
\end{tabular}
\\
\bottomrule
\end{tabular}
\vspace{2pt}
\caption{
\textbf{List of prompts used for evaluation.} For seen-object evaluations, \slot{descriptor} refers to objects observed during training, i.e., ``black'' or ``plastic.'' For unseen-object evaluations, it refers to novel descriptors such as ``brown'' or ``red.'' In the novel-language setting, \slot{value} denotes ordinal identifiers, i.e., 1, 2, or 3, used to evaluate compositional reference grounding under previously unseen instruction patterns.
}
\label{tab:prompt_list}
\end{table}

\begin{figure}[h]
    \centering
    % \vspace{-6pt}
    \includegraphics[width=0.95\linewidth]{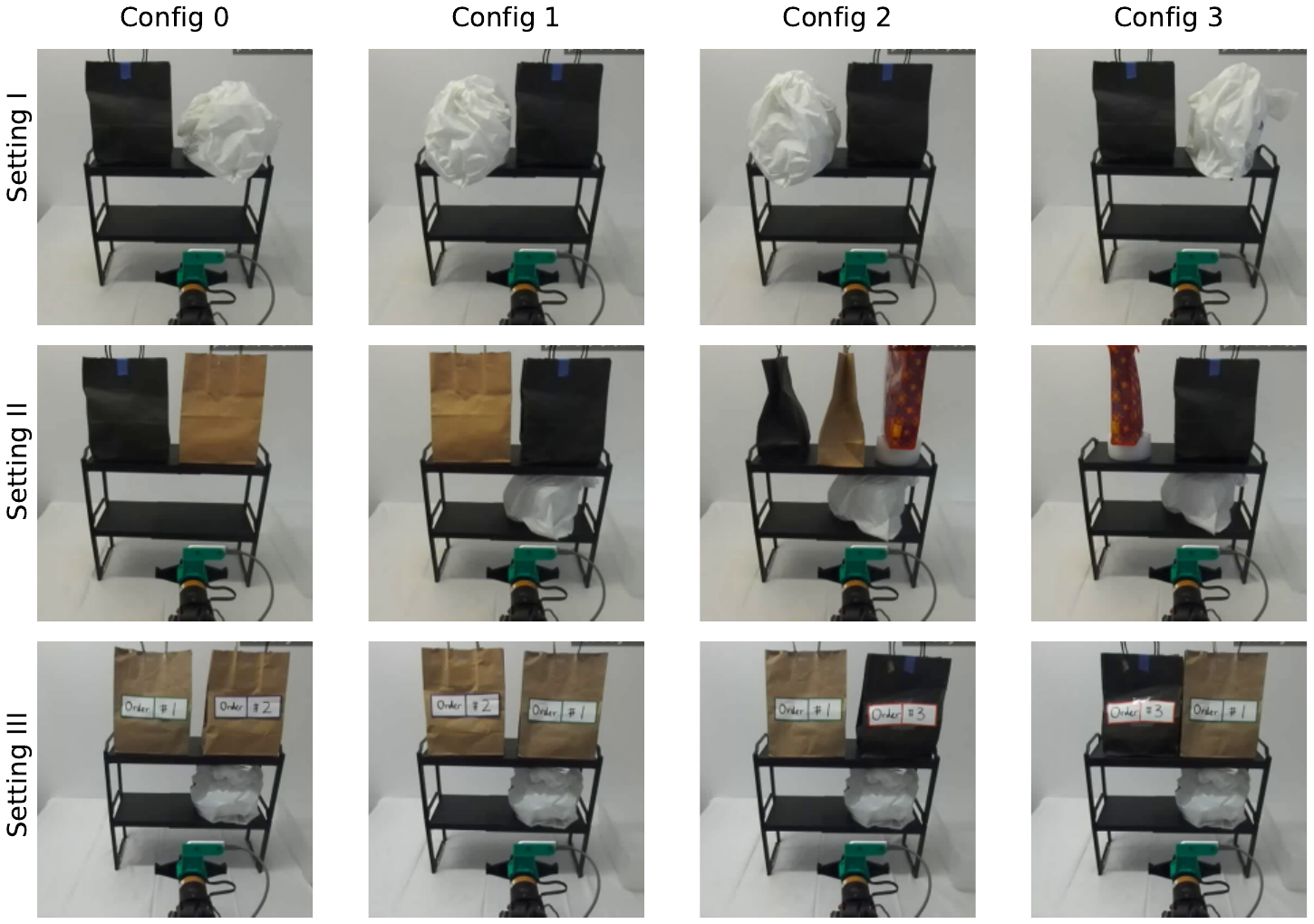}
    \vspace{-6pt}
    \caption{
        \textbf{Representative evaluation configurations across the three experimental settings.} Rows correspond to the evaluation regimes described in the main paper: Setting~I uses seen objects with paraphrased language, Setting~II uses unseen objects with known language, and Setting~III uses unseen objects with novel language. Columns show different spatial layouts used during evaluation, varying object placement, distractor arrangement, and scene geometry. In Setting~I, Configurations~0--1 correspond to the black-bag task, while Configurations~2--3 correspond to the plastic-bag task. All methods are evaluated on the same prompts and spatial configurations.
    }
    \label{fig:setting_configurations}
    \vspace{-8pt}
\end{figure}

In addition to language variation, we evaluate robustness under diverse spatial
configurations and object layouts. Figure~\ref{fig:setting_configurations}
shows representative scenes from each evaluation setting. Across configurations, we vary object placement, relative spacing, distractor positions, and camera viewpoints while preserving the same high-level pick-and-place objective. The seen-object setting reuses training objects in novel arrangements, whereas the unseen-object settings introduce novel targets and increasingly challenging referring expressions. All configurations are held fixed across compared methods to ensure consistent evaluation conditions.

\section{Additional Results}
\begin{figure}[h]
    \centering
    % \vspace{-6pt}
    \includegraphics[width=1.0\linewidth, trim=170 130 170 60, clip]{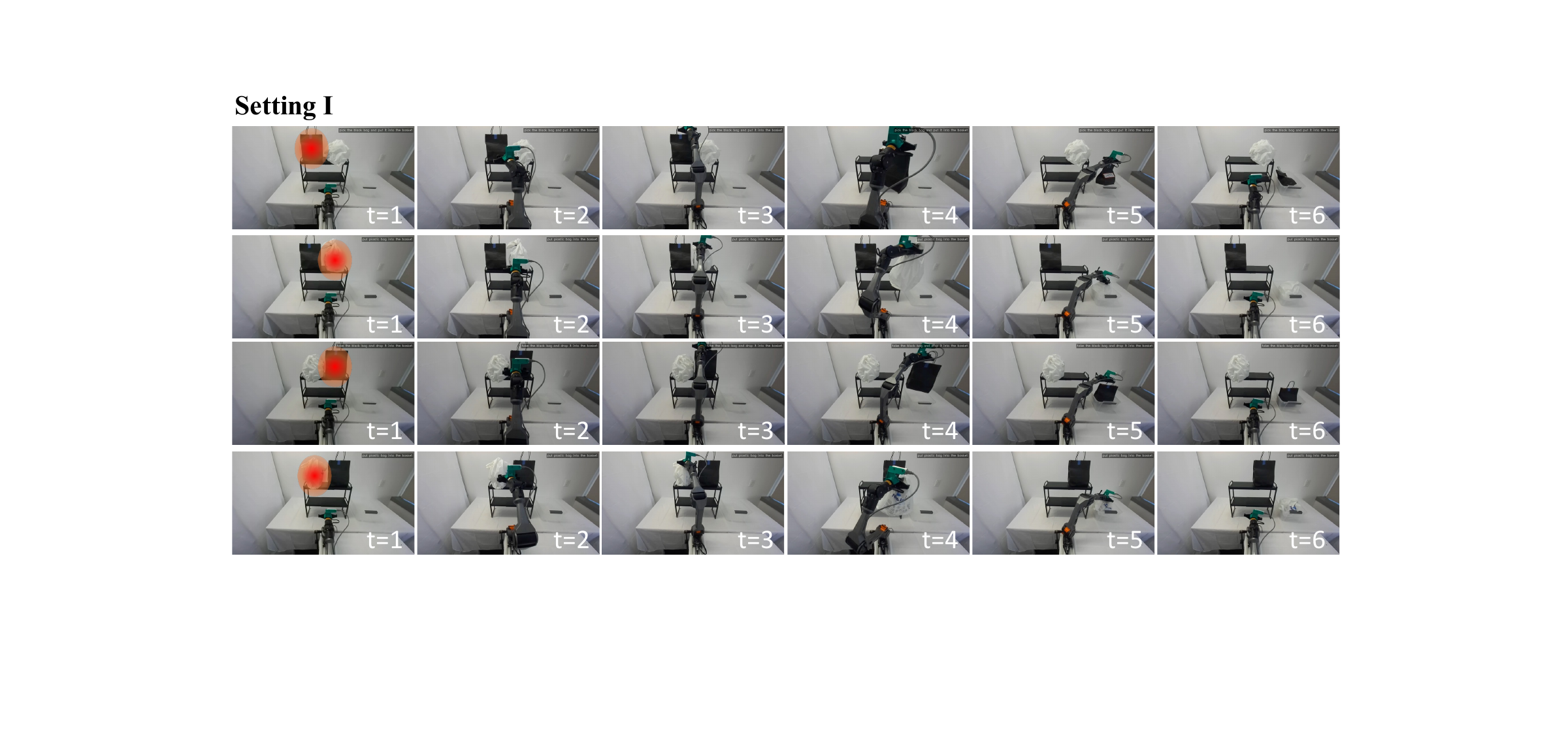} \\ 
    \includegraphics[width=1.0\linewidth, trim=170 130 170 60, clip]{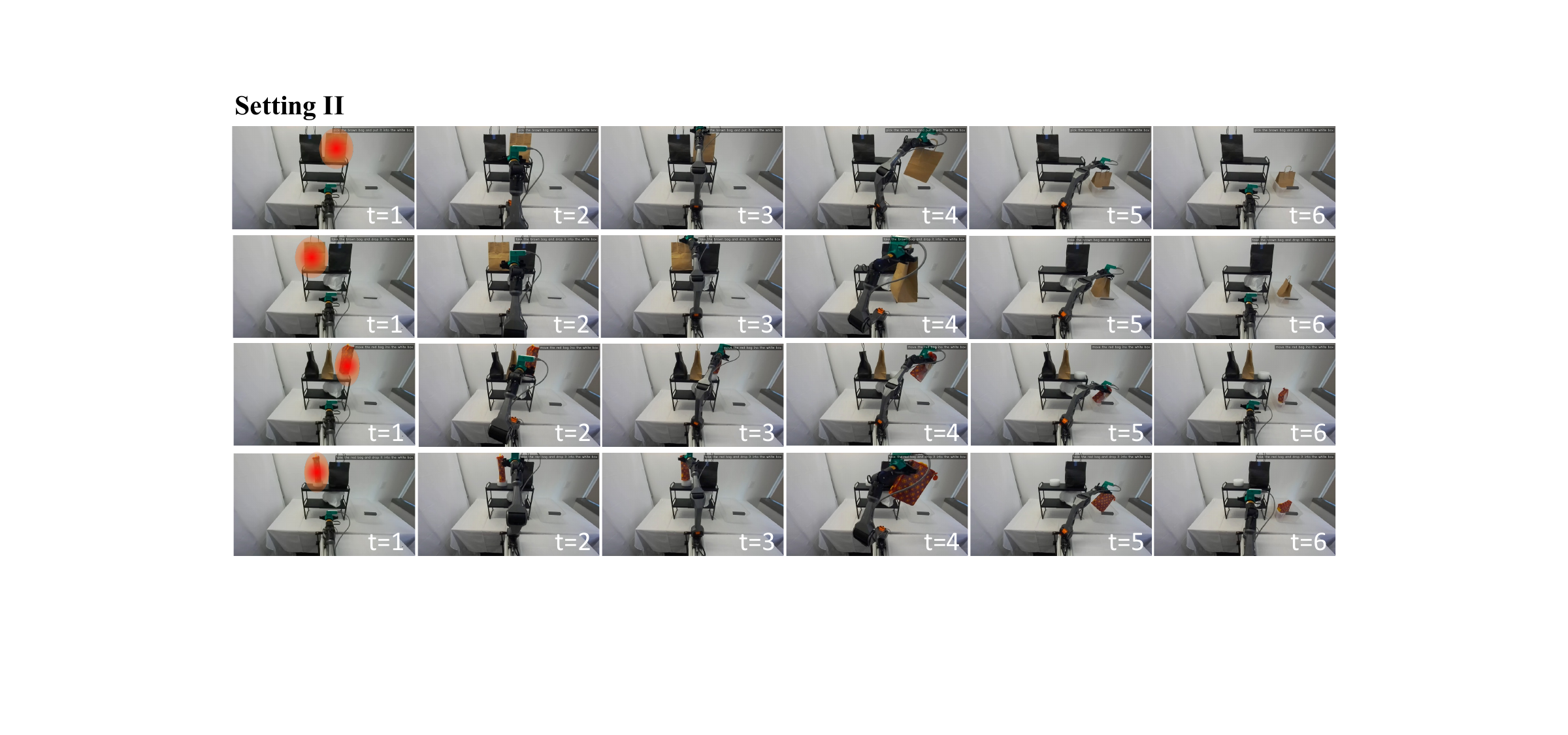} \\
    \includegraphics[width=1.0\linewidth, trim=170 130 170 60, clip]{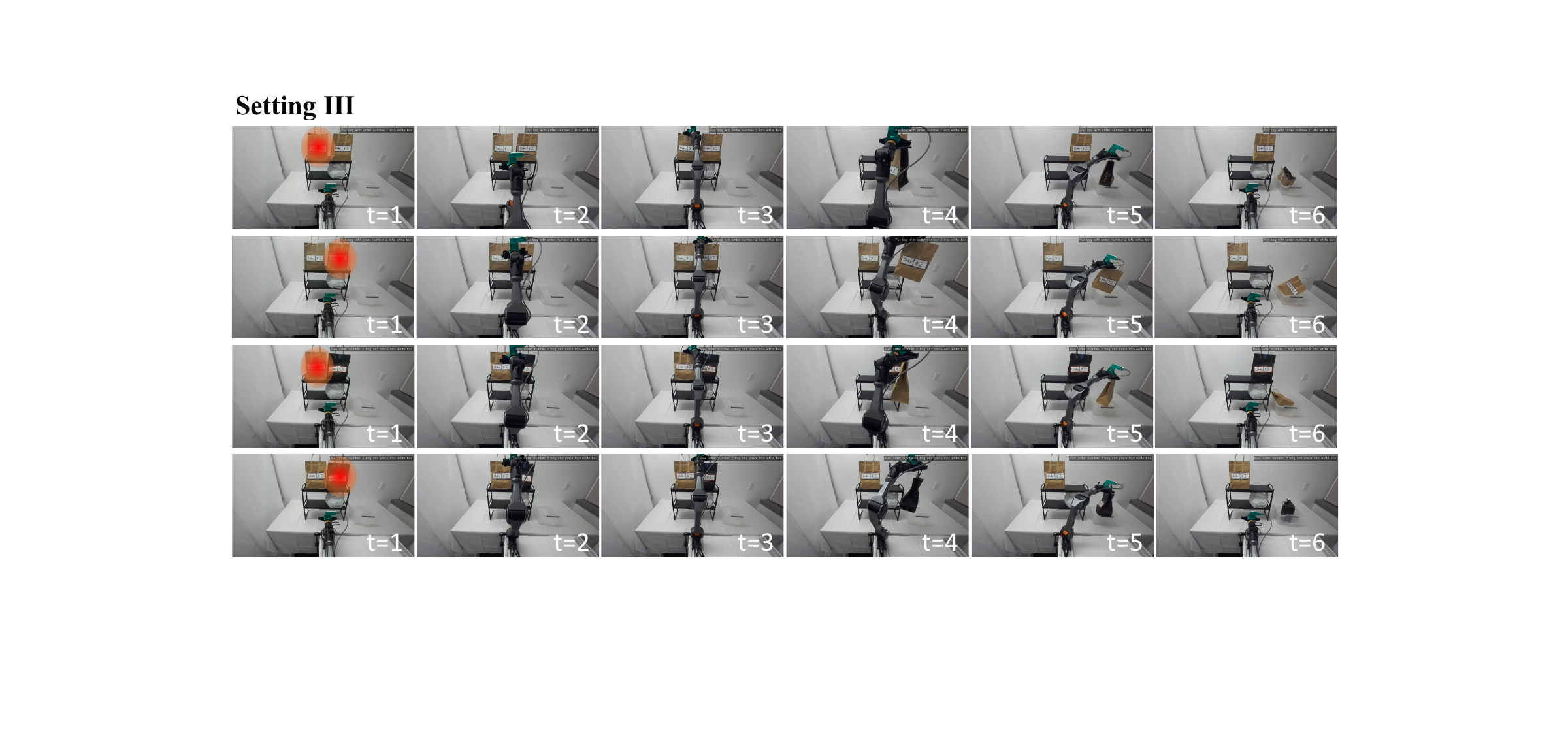}
    \vspace{-16pt}
    \caption{\textbf{Example demonstrations of policy steering across the three evaluation settings.} The red heatmap indicates the correctly grounded target object.}
    \label{fig:steer_appendix}
    \vspace{-8pt}
\end{figure}

Figure~\ref{fig:steer_appendix} provides additional visual comparisons, where the red heatmap indicates the correctly grounded target object. Under the same spatial object layouts, our method consistently steers the policy toward the spatial cue across all three evaluation settings. We observe that the policy often produces similar actions in the early execution steps, but as the end-effector approaches the objects, the generated trajectory is progressively redirected toward the cued target.

\section{Failure Cases}
Our method can fail in two scenarios. First, it cannot recover when the underlying manipulation prior is unreliable. Noisy rollouts, accumulated execution errors, or environmental changes can still cause the frozen policy to deviate from a feasible manipulation trajectory. Second, the steering strength is applied consistently throughout the rollout, which can over-modulate the policy output when precise actions are required. This may lead to manipulation errors such as overshooting or unstable grasping.

% \section{Video Demonstrations}
% The supplementary material includes a paper overview video and the corresponding full policy rollout demonstrations.

\end{document}